\documentclass[journal]{IEEEtran}
\usepackage{graphicx}

\usepackage{amsmath}
\usepackage{amssymb}
\usepackage{amsfonts}

\usepackage{booktabs}
\usepackage{multirow}
\usepackage{subfig}
\usepackage{cite}
\usepackage{floatflt}
\usepackage{wrapfig}
\usepackage[linesnumbered, ruled,vlined]{algorithm2e}
\usepackage{url}

\def\vector#1{\mbox{\boldmath $#1$}}
\newcommand{\CR}{C}

\newtheorem{observation}{Observation} 

\ifCLASSINFOpdf
\else
\fi
\begin{document}
%


\title{Reviewing and Benchmarking Parameter Control Methods in Differential Evolution}

%
%
%

\author{
Ryoji~Tanabe,~\IEEEmembership{Member,~IEEE,}and~Alex~Fukunaga 
\thanks{R. Tanabe is with the Shenzhen Key Laboratory of Computational Intelligence, University Key Laboratory of Evolving Intelligent Systems, Guangdong Province, Department of Computer Science and Engineering, Southern University of Science and Technology, Shenzhen 518055, China. e-mail: (rt.ryoji.tanabe@gmail.com).}
\thanks{A. Fukunaga is with the Graduate School of Arts and Sciences, The University of Tokyo, 3-8-1 Komaba, Meguro-ku, Tokyo, Japan. e-mail: (fukunaga@idea.c.u-tokyo.ac.jp).}
}

\maketitle


\begin{abstract}

Many Differential Evolution (DE) algorithms with various parameter control methods (PCMs) have been proposed.
However, previous studies usually considered PCMs to be an integral component of a complex DE algorithm. Thus the characteristics and performance of each method are poorly understood.
We present an in-depth review of 24 PCMs for the scale factor and crossover rate in DE and a large scale benchmarking study.
We carefully extract the 24 PCMs from their original, complex algorithms and describe them according to a systematic manner.
Our review facilitates the understanding of similarities and differences between existing, representative PCMs.
The performance of DEs with the 24 PCMs and 16 variation operators is investigated on 24 black-box benchmark functions.
Our benchmarking results reveal which methods exhibit high performance when embedded in a standardized framework under 16 different conditions, independent from their original, complex algorithms.  
We also investigate how much room there is for further improvement  of PCMs by comparing the 24 methods with an oracle-based model, which can be considered to be a conservative lower bound on the performance of an optimal method.

\end{abstract}

\begin{IEEEkeywords}
Continuous optimization, parameter control methods, differential evolution, review, benchmarking study
\end{IEEEkeywords}

%
\IEEEpeerreviewmaketitle

\section{Introduction}
\label{sec:introduction}





\IEEEPARstart{T}{he} performance of Evolutionary Algorithms (EAs) is significantly influenced by control parameter settings \cite{EibenHM99,KarafotiasHE15}.
The optimal control parameter settings depend on the characteristics of the problem instance which is being solved, and it is necessary to tune the control parameters to obtain the good performance of an EA for a given problem instance.
Suitable control parameter settings are also dependent on the state of the search progress of an EA.
Thus, Parameter Control Methods (PCMs), which are methods for automatically tuning control parameters during the search process, have been widely studied \cite{EibenHM99, KarafotiasHE15}.
According to \cite{EibenHM99}, PCMs can be classified into the following three broad categories:

(1) {\em Deterministic Parameter Control Methods} (DPCMs) which set control parameter values according to a deterministic rule without using any information obtained during the search process.
DPCMs include, for example, a method that decreases mutation rates according to the computational resources used so far (the number of function evaluations or runtime) \cite{Fogarty89}.


(2) {\em Adaptive Parameter Control Methods} (APCMs) which set global control parameters according to the current state of the search.
To our knowledge, the earliest APCM is the ``$1/5$ success rule'' for Evolution Strategy (ES), which sets the step size parameter $\sigma$ according to a simple adaptive criterion \cite{Rechenberg73}.

(3) {\em Self-adaptive Parameter Control Methods} (SPCMs), in which each individual in a population has its own associated control parameter values.
These individual-specific control parameter values are encoded as part of the individual, and thus the control parameters themselves are subject to selection as well as modification by genetic operators, with the aim of simultaneously evolving the individuals as well as the individual-specific control parameters. 
Representative instances of this approach include Self-adaptive ES \cite{BackS93}. 
The key difference between APCMs and SPCMs is whether new control parameter values are generated by applying variation operators to values associated with each individual.

Differential Evolution (DE) is an EA that was
primarily designed for continuous optimization problems \cite{StornP97}.
Continuous optimization is the problem of finding a real-valued vector $\vector{x} = (x_1, ..., x_D)^{\rm T} \in \mathbb{S} \subseteq \mathbb{R}^D$ 
 that minimizes an objective function $f: \mathbb{S} \rightarrow \mathbb{R}$, where $D$ is the dimensionality of the problem, and $\mathbb{S}$ is the feasible region of the search space.
Despite its relative simplicity, DE has been shown to be competitive with more complex optimization algorithms and
has been applied to many practical problems \cite{PriceSL05,DasS11,DasMS16}.


The fact that the performance of EAs is strongly influenced by its control parameter settings was widely accepted in the evolutionary computation community for decades.
Nevertheless, it was initially reported that the performance of DE was fairly robust with respect to control parameter settings \cite{StornP97}.
However, later research showed that in fact, the effectiveness of DE was significantly affected by control parameter settings \cite{GamperleMK02, ZielinskiWLK06, BrestGBMZ06}.
The main control parameters of DE are the population size $N$, the scale parameter $F >0$, and the crossover rate $\CR \in [0,1]$.
As a result, PCMs for DE have been an active area of research since around 2005. 
Many DE algorithms with PCMs have been proposed \cite{NeriT10,DasS11,DasMS16,DragoiD16}.

In general, the term ``a DE algorithm with a PCM'' denotes a complex algorithm composed of multiple components.
For example, ``L-SHADE'' \cite{TanabeF14CEC} consists of four key components: (a) current-to-$p$best/1 mutation strategy \cite{ZhangS09}, (b) binomial crossover, (c) the ``SHADE method'' for adapting  parameters  $F$ and $\CR$ (i.e., PCM-SHADE), and (d) linear population size reduction strategy. 
In the same manner, ``CoDE'' \cite{WangCZ11} consists of four components: (a) three mutation strategies (rand/1, rand/2, and current-to-rand/1), (b) binomial crossover, (c) the ``CoDE method'' for assigning $F$ and $\CR$ values to each individual (i.e., PCM-CoDE), and (d) simultaneous generation of three children per an individual.





While there have been a number of performance comparisons among such complex DE algorithms, recent work \cite{ZielinskiWL08, DrozdikAAT15} pointed out that there are few comparative studies of PCMs \emph{in isolation}.
For example, a previous study \cite{TanabeF14CEC} compares L-SHADE with five complex DE algorithms with PCMs (SaDE \cite{QinHS09}, dynNP-jDE \cite{BrestM08}, JADE \cite{ZhangS09}, EPSDE \cite{MallipeddiSPT11}, and CoDE \cite{WangCZ11}), but a performance comparison between their PCMs (PCM-SHADE, PCM-SaDE, PCM-jDE, PCM-JADE, PCM-EPSDE, and PCM-CoDE) has never been performed in the literature.
The results in \cite{TanabeF14CEC} show that L-SHADE performs better than the five DE algorithms.
However, the results do not mean that the performance of PCM-SHADE is better than that of the five PCMs, because it is unclear which component ((a)-(d) in the above example) makes the largest contribution to the better performance of L-SHADE.
Thus, the performance of (c) PCMs is currently poorly understood -- a deeper understanding is necessary to enable the development of new PCMs and further advance the state of the art.


In this paper, we first present an \emph{in-depth} review of 24 PCMs in DE which yields novel insights.
Existing survey papers (e.g., \cite{NeriT10,DasS11,ChiangCL13,KarafotiasHE15,DasMS16,DragoiD16}) have a much broader scope, covering all aspects of DE. 
\emph{In contrast, this paper focuses only on PCMs for the scale factor $F$ and crossover rate $\CR$}.
We extracted only (c) PCMs from complex DE algorithms and generalized them so that they can be combined with arbitrary mutation strategies and crossover methods.
Instead of an exhaustive, shallow survey of PCMs,
we precisely describe the 24 PCMs and clarify their relationship to each other.

Next, we present a large-scale benchmarking study which investigates the performance of the 24 PCMs on the noiseless BBOB benchmark set \cite{hansen2012fun}.
Each PCM is incorporated into DE algorithms with eight mutation strategies and two crossover methods.
By fixing the (a) mutation strategy and (b) crossover method for all PCMs, the performance of (c) PCMs can be investigated independently.
Furthermore, we evaluate the performance of the 24 PCMs with hyperparameters tuned by SMAC \cite{HutterHL11}, which is an automatic algorithm configurator.

Finally, the 24 PCMs are compared with GAODE \cite{TanabeF16}, a recently proposed, oracle-based model of an ``ideal PCM'', which can be considered as a lower bound on the performance of PCMs in DE.
This shows that the performance of existing PCMs is poor compared with an (approximately) ideal PCM, and indicates that there is significant room for improvement and opportunities for  future work in the development of PCMs.

This paper is the first study which reviews many PCMs for DE and performs a systematic investigation their performance in isolation.
While there have been some benchmarking studies of DE algorithms with PCMs (e.g., \cite{BrestBGZM07,Tvrdik09}) as well as survey papers \cite{NeriT10,DasS11,ChiangCL13,DasMS16,DragoiD16}, they did not focus specifically on  PCMs.
Several previous studies have compared the performance of PCMs (e.g., \cite{ZielinskiWL08,DrozdikAAT15,SeguraCSL15,SeguraCSL14}).
However, the benchmark problems used in the previous studies \cite{ZielinskiWL08} and \cite{DrozdikAAT15} were constrained continuous optimization and multi-objective optimization problems.
Thus, the performance of PCMs on bound-constrained continuous optimization problems, the most basic class of optimization problems as represented by the BBOB benchmarks, has not been thoroughly investigated.
Also, the previous studies \cite{SeguraCSL15,SeguraCSL14} investigate adaptation mechanisms only for either the scale factor or crossover rate, but not both.
Most importantly, all of the previous work \cite{ZielinskiWL08,DrozdikAAT15,SeguraCSL15,SeguraCSL14} evaluated the performance of only a few PCMs, and only considered up to two combinations of mutation and crossover methods.

In most previous studies, PCMs have been considered integral, algorithmic components of complex DE algorithms.
Thus, the similarities and differences between existing PCMs were often not obvious.
Our review classifies and organizes existing PCMs by presenting a detailed, precise description of each method in a common framework. 
This should facilitate the understanding of the characteristics of each PCM as well as the relationship among the various methods.
Furthermore, there has a recent tendency for ``horse-race'' algorithm development in the DE community -- many DE algorithms that include new PCMs embedded in a complex DE have been proposed, where the performance of the entire algorithm is evaluated without isolating the performance impact of the PCMs.
Our benchmarking study aims to reveal which PCM tend to perform when embedded in a standard DE framework, rather than which complex DE algorithm has a good performance.
Source code for the 24 PCMs as well as all experimental data presented in this paper are available online (\url{https://sites.google.com/view/pcmde/}) so that researchers can easily compare their newly developed PCMs with the 24 PCMs.
We hope that our review and benchmarking results encourage further development of PCMs for DE.



This paper is organized as follows:
First, Section \ref{sec:de} describes the basic DE framework.
Then, Section \ref{sec:pcm} reviews PCMs proposed in the literature.
The historical background of PCMs, their characteristics, and the relationships between different PCMs are discussed in Section \ref{sec:discussion-pcms}.
The benchmarking results are reported in Section \ref{sec:experimental-results}.
Finally, Section \ref{sec:conclusion} concludes this paper and discusses directions for future work.

\section{Differential Evolution}
\label{sec:de}

This section briefly describes the basic DE algorithm \cite{StornP97} without any PCM (i.e., the $F$ and $\CR$ parameters are fixed values, such as $F=0.5$ and $\CR = 0.9$).
A DE population $\vector{P} = \{ \vector{x}^{1}, ..., \vector{x}^{N} \}$ is represented as a set of real parameter vectors $\vector{x}^{i} = (x^{i}_{1}, ..., x^{i}_{D})^\mathrm{T}$, $i \in \{1, ..., N\}$, where $D$ is the dimensionality of a given problem, and $N$ is the population size.
The detailed pseudocode  for the DE described here can be found in Algorithm S.1 in the supplementary file. 


After the initialization of the population, for each iteration $t$, for each $\vector{x}^{i,t}$, a mutant vector $\vector{v}^{i,t}$ is generated from the individuals in $\vector{P}^t$ by applying a mutation strategy.
Table S.1 in the supplementary file shows eight representative mutation strategies for DE (rand/1, rand/2, best/1, best/2, current-to-rand/1, current-to-best/1 \cite{ZhangS09}, current-to-$p$best/1, and rand-to-$p$best/1 \cite{ZhangS09_books}).
The scale factor $F$ in Table S.1 controls the magnitude of the differential mutation.


Then, the mutant vector $\vector{v}^{i,t}$ is crossed with the parent $\vector{x}^{i,t}$ to generate a trial vector $\vector{u}^{i,t}$.
Binomial crossover (bin) and exponential crossover (exp) are most commonly used in DE.
However, the performance of a DE algorithm using exponential crossover significantly depends on the ordering of variable indices \cite{TanabeF14PPSN}.
Shuffled exponential crossover (sec), which applies exponential crossover after shuffling the variable indices of parent individuals, addresses this issue \cite{PriceSL05,TanabeF14PPSN}.
Algorithms S.2 -- S.4 in the supplementary file show the three crossover methods, respectively.
The crossover rate $\CR$ in Algorithms S.2 -- S.4 controls the number of inherited variables from  $\vector{x}^{i,t}$ to $\vector{u}^{i,t}$.


After all of the trial vectors have been generated, for each $i$, $\vector{x}^{i,t}$ is compared with its corresponding $\vector{u}^{i,t}$, keeping the better individual in the population $\vector{P}^{t+1}$. 
The parent individuals that were worse than the trial vectors (and are therefore not selected for survival in the standard DE) are preserved in an external archive $\vector{A}$.
Whenever the size of the archive exceeds $|\vector{A}|$, randomly selected individuals are deleted to keep the archive size constant.
Individuals in $\vector{A}$ are used for the current-to-$p$best/1 and rand-to-$p$best/1 mutation strategies.

\section{Parameter Control Methods in DE}
\label{sec:pcm}


As mentioned in Section \ref{sec:introduction}, PCMs can be classified into the following three categories: DPCMs, APCMs, and SPCMs.
This section reviews 24 PCMs for DE in each category in Sections \ref{sec:dpcm}, \ref{sec:apcm}, and \ref{sec:spcm}, respectively.
The properties of the 24 PCMs are summarized in Table \ref{tab:pcms}, which is explained in Section \ref{sec:discussion-pcms}.
For the sake of clarity, MDE$\_p$BX \cite{IslamDGRS12}, MDE \cite{ZouWGL13}, and ADE  \cite{YuSCZGLLZ14} are denoted as IMDE, ZMDE, and YADE, respectively.
To improve understandability, we describe PCM-jDE \cite{BrestGBMZ06}, PCM-FDSADE \cite{TirronenN09}, and PCM-ISADE \cite{JiaGW09} in a revised manner from their original description.
Some papers provide insufficient details on the proposed PCMs to allow reimplementation, 
so we excluded such insufficiently specified PCMs from this paper.

In \cite{ZhangS09}, Zhang and Sanderson point out that some APCMs are incorrectly classified by the original authors as SPCMs with respect to the taxonomy proposed in \cite{EibenHM99}.
For example, Self-adaptive DE (SaDE) \cite{QinHS09} does not use any variation operators to generate the values of $\CR$. Thus, even though the name ``SaDE'' includes the term ``Self-adaptive'', the PCM of SaDE should be classified as an APCM.
Following Zhang and Sanderson's guideline, we have classified PCMs according to their functional mechanisms, rather than their original names.


Algorithms S.5--S.28 in the supplementary file show a basic DE algorithm (see Section \ref{sec:de}) with each generalized PCM, respectively.
Note that the pseudocode in Algorithms S.5--S.28 describes the minimally modified DE algorithms to incorporate into each PCM but does not represent the exact DE algorithms described in the original paper.
It should be re-emphasized that in this paper, we survey and evaluate PCMs, which are the parameter control components used in previous DE algorithms (e.g., PCM-DERSF \cite{DasKC05a}), and not the complex DE algorithms that consist of multiple components (e.g., DERSF).
See the original papers for full descriptions of the entire, complex, original algorithms which incorporates the 24 PCMs.


On the one hand, the $F_{i,t}$ and $\CR_{i,t}$ represent values of $F$ and $\CR$ assigned to an individual $\vector{x}_{i,t}$ ($i \in \{1, ..., N\}$) in the population $\vector{P}^t$ at iteration $t$ (i.e., each individual uses different values of $F$ and $\CR$).
On the other hand, $F_{t}$ and $\CR_{t}$ represent values of $F$ and $\CR$ assigned to all individuals in $\vector{P}^t$ (i.e., all individuals use same values of $F$ and $\CR$).
The function ${\rm randu} [a,b]$ denotes a uniformly selected random number from $[a, b]$, and ${\rm randn}(\mu, \sigma^2)$ is a value selected randomly from a Normal distribution with mean $\mu$ and variance $\sigma^2$.
The function ${\rm randc}(\mu, \sigma)$ selects values randomly from a Cauchy distribution with location parameter $\mu$ and scale parameter $\sigma$.
Unless explicitly noted, when values of $F$ and $\CR$ generated by a PCM are outside of $[0, 1]$, they are replaced by the closest, limit value (0 or 1) .
We say that a generation of a trial vector is {\em successful} if $ f(\vector{u}^{i,t}) \leq f(\vector{x}^{i,t})$.
Otherwise, we say that the trial vector generation is a {\em failure}.

\subsection{DPCMs in DE}
\label{sec:dpcm}


\noindent {\bf $\bullet$ PCM-DERSF} (Algorithm S.5):
The PCM of DE with Random Scale Factor (PCM-DERSF) \cite{DasKC05a} uniformly randomly generates the value of $F_{i,t}$ in the range $[F^{\rm min}, F^{\rm max}]$ as follows: $F_{i,t}  = {\rm randu}[F^{\rm min}, F^{\rm max}]$.
The recommended values of $F^{\rm min}$ and $F^{\rm max}$ are $0.5$ and $1$, respectively. 
It is possible to generate diversified mutant individuals by randomly generating the $F$ parameter.
While the $F$ values are randomly generated, $\CR$ is a constant value ($\CR = 0.9$ is recommended).

\noindent {\bf $\bullet$ PCM-DETVSF} (Algorithm S.6):
In the PCM of DE with Time Varying Scale Factor (PCM-DETVSF) \cite{DasKC05a}, the $F$ value decreases linearly with the number of iterations $t$ as follows: $F_{t}  = (F^{\rm max}-F^{\rm min}) \bigl( \frac{t^{\rm max} - t}{t^{\rm max}} \bigr) + F^{\rm min}$, 
%
where $t^{\rm max}$ is the maximum number of iterations, and 
$F_1$ and $F_{t^{\rm max}}$ are equal to $F^{\rm max}$ and $F^{\rm min}$, respectively.
The recommended settings of $F^{\rm min}$ and $F^{\rm max}$ are $0.4$ and $1.2$, respectively.
These settings are based on a general rule of thumb that the explorative and exploitative searches should be performed at the beginning and end of the search process.
Larger and smaller $F$ values are appropriate for each purpose.
Similar to PCM-DERSF, $\CR$ is a constant value ($\CR = 0.9$).


\noindent {\bf $\bullet$ PCM-SinDE} (Algorithm S.7):
The PCM of Sinusoidal DE (PCM-SinDE) \cite{DraaBB15} determines the values of $F$ and $\CR$ based on the sinusoidal function as follows:
{\small
\begin{align}
\label{eqn:sindef}
\small
F_{t}  &= \frac{1}{2} \biggl( \frac{t}{t^{\rm max}} \bigl({\rm sin}(2 \pi \omega t)\bigr) + 1 \biggr),\\
\label{eqn:sindecr}
\CR_{t}  &= \frac{1}{2} \biggl( \frac{t}{t^{\rm max}} \bigl({\rm sin}(2 \pi \omega t + \pi)\bigr) + 1 \biggr),
\end{align}
}%
\noindent where $\omega$ is the angular frequency, and a value of  $\omega = 0.25$ is recommended.
The amplitude of the $F$ and $\CR$ values depends on the number of iterations $t$.
Thus, the ranges of the possible values of $F$ and $\CR$ increase as the search progresses.


\noindent {\bf $\bullet$ PCM-ZMDE} (Algorithm S.8):
The PCM of Zou's Modified DE (PCM-ZMDE) \cite{ZouWGL13} randomly generates the $F$ and $\CR$ values according to the Normal and uniform distributions as follows: $F_{i,t}  = {\rm randn}(0.75, 0.1)$ and $\CR_{i,t}  = {\rm randu}[0.8, 1]$.
%

\noindent {\bf $\bullet$ PCM-CoDE} (Algorithm S.9):
The PCM of Composite DE (PCM-CoDE) \cite{WangCZ11} uses three pre-defined pairs of $F$ and $\CR$ values for parameter control: 
$\vector{q}^1 = (1, 0.1)$,  $\vector{q}^2 = (1, 0.9)$,  and $\vector{q}^3 = (0.8, 0.2)$, where  $\vector{q} = (F, \CR)$.
These combinations were determined based on frequently used parameter settings in the DE community.
At the beginning of each iteration $t$, a randomly selected pair $\vector{q}$ is assigned to each individual.


\noindent {\bf $\bullet$ PCM-SWDE} (Algorithm S.10):
At the beginning of each iteration $t$, the PCM of Switching DE (PCM-SWDE) \cite{DasGM15} randomly assigns the extreme values of $F$ ($0.5$ or $2$) and $\CR$ ($0$ or $1$) to each individual.
The small and large $F$ values are helpful for the explorative and the exploitative searches, respectively.
The use of the two extreme $\CR$ values is to strike a balance between variable-wise and vector-wise searches.

\subsection{APCMs in DE}
\label{sec:apcm}

\noindent {\bf $\bullet$ PCM-DEPD} (Algorithm S.11):
In the PCM of DE using Pre-calculated Differential (PCM-DEPD) \cite{AliT04}, only the $F$ parameter is adaptively adjusted based on the objective values of individuals in the population $\vector{P}^t$.
For each iteration $t$, $F_t$ is given as follows:
{\small
\begin{align}  
  \label{eqn:atde}
F_t = \begin{cases}
{\rm max}\{F^{\rm min}, 1-|\frac{f^{\rm max}_t}{f^{\rm min}_t}| \} &   {\rm if} \: |\frac{f^{\rm max}_t}{f^{\rm min}_t}| < 1\\
{\rm max}\{F^{\rm min}, 1-|\frac{f^{\rm min}_t}{f^{\rm max}_t}| \} &   {\rm otherwise}
  \end{cases},
\end{align}
}%
where $f^{{\rm max}}_t$ and $f^{{\rm min}}_t$ are the maximum and minimum objective values in $\vector{P}^t$.
The recommended value for $F^{\rm min}$ is $0.4$.
Unlike $F$, $\CR$ is a fixed value ($\CR=0.5$).

\noindent {\bf $\bullet$ PCM-jDE} (Algorithm S.12):
The PCM of Janez's DE (PCM-jDE) \cite{BrestGBMZ06} assigns a different set of parameter values $F_{i,t}$ and $\CR_{i,t}$ to each $\vector{x}^{i,t}$ in $\vector{P}^t$.
For $t=1$, the parameters for all individuals $\vector{x}^{i,t}$ are set to $F_{i,t} = 0.5$ and $\CR_{i,t} = 0.9$.
In each iteration $t$, $F^{\rm trial}_{i,t}$ and $\CR^{\rm trial}_{i,t}$ are generated as follows:
{\small
\begin{align}
\label{eqn:jde-f}
F^{\rm trial}_{i,t} &= \begin{cases}
{\rm randu}[0.1,1] &   {\rm if} \: {\rm randu}[0,1] < \tau_F\\
F_{i,t} &   {\rm otherwise}
  \end{cases},
\\
\label{eqn:jde-cr}
\small
\CR^{\rm trial}_{i,t} &= \begin{cases}
{\rm randu}[0,1]  &   {\rm if} \: {\rm randu}[0,1] < \tau_{\CR}\\
\CR_{i,t} &   {\rm otherwise}
  \end{cases},
\end{align}
}%
where $\tau_F$ and $\tau_{\CR} \in (0,1]$ are hyperparameters for parameter adaptation ($\tau_F = \tau_{\CR} = 0.1$ is the recommended setting). 
For each individual, $\vector{u}^{i,t}$ is generated by using $F^{\rm trial}_{i,t}$ and $\CR^{\rm trial}_{i,t}$ in \eqref{eqn:jde-f} and \eqref{eqn:jde-cr}.
When the trial is a success, each individual $\vector{x}^{i,t}$ uses $F^{\rm trial}_{i,t}$ and $\CR^{\rm trial}_{i,t}$ in the next iteration (i.e., $F_{i,t+1} = F^{\rm trial}_{i,t}$ and $\CR_{i,t+1} = \CR^{\rm trial}_{i,t}$).
Otherwise, each individual keeps using $F_{i,t}$ and $\CR_{i,t}$  (i.e., $F_{i,t+1} = F_{i,t}$ and $\CR_{i,t+1} = \CR_{i,t}$).

\noindent {\bf $\bullet$ PCM-FDSADE} (Algorithm S.13):
The PCM of Fitness Diversity Self-Adaptive DE (PCM-FDSADE) \cite{TirronenN09} is a variant of PCM-jDE.
Similar to PCM-jDE, PCM-FDSADE generates $F^{\rm trial}_{i,t}$ and $\CR^{\rm trial}_{i,t}$ for each individual and updates $F_{i,t+1}$ and $\CR_{i,t+1}$ based on the success/failure decision.
However, PCM-FDSADE samples $F^{\rm trial}_{i,t}$ and $\CR^{\rm trial}_{i,t}$ based on the diversity of the objective values of individuals as follows:
%
%
{\small
\begin{align}
\label{eqn:fdsade-f}
\small
F^{\rm trial}_{i,t} &= \begin{cases}
{\rm randu}[0.1,1] &   {\rm if} \: {\rm randu}[0,1] < K (1 - \phi_t)\\
F_{i,t} &   {\rm otherwise}
  \end{cases},
\\
\label{eqn:fdsade-cr}
\small
\CR^{\rm trial}_{i,t} &= \begin{cases}
{\rm randu}[0,1]  &   {\rm if} \: {\rm randu}[0,1] < K (1 - \phi_t)\\
\CR_{i,t} &   {\rm otherwise}
  \end{cases},
\end{align}
}%
where $K$ is the control parameter of PCM-FDSADE ($K=0.3$ is recommended).
The value of $\phi_t$ indicates the diversity of the population $\vector{P}^t$ in the objective space, where $\phi_t = f^{\rm std}_t / (f^{\rm max}_t - f^{\rm min}_t)$, 
$f^{\rm std}_t$ is the standard deviation of $f(\vector{x}^{1,t}), ..., f(\vector{x}^{N,t})$, 
and $f^{\rm max}_t$ and $f^{\rm min}_t$ are their maximum and minimum objective values, respectively.
If $f^{\rm max}_t - f^{\rm min}_t < 0$, $\phi_t=0$.

\noindent {\bf $\bullet$ PCM-ISADE} (Algorithm S.14):
The PCM of Improved Self-Adaptive DE (PCM-ISADE) \cite{JiaGW09} is also based on PCM-jDE, but $F^{\rm trial}_{i,t}$ and $\CR^{\rm trial}_{i,t}$ are generated based on the objective values of individuals $f(\vector{x}^{1,t}), ..., f(\vector{x}^{N,t})$ as follows:
%
%
{\footnotesize
\begin{align}
\label{eqn:isade-f}
%
F^{\rm trial}_{i,t} &= \begin{cases}
\alpha (F_{i,t} - 0.1) + 0.1 & {\rm if} \: {\rm randu}[0,1] < \tau_F \: \& \: f(\vector{x}^{i,t}) < f^{{\rm avg}}_t\\
{\rm randu}[0.1,1] & {\rm if} \: {\rm randu}[0,1] < \tau_F \: \& \: f(\vector{x}^{i,t}) \geq f^{{\rm avg}}_t\\
F_{i,t} &   {\rm otherwise}
\end{cases},
\\
\label{eqn:isade-cr}
\CR^{\rm trial}_{i,t} &= \begin{cases}
\alpha \, \CR_{i,t} & {\rm if} \: {\rm randu}[0,1] < \tau_{\CR} \: \& \: f(\vector{x}^{i,t}) < f^{{\rm avg}}_t\\
{\rm randu}[0,1] & {\rm if} \: {\rm randu}[0,1] < \tau_{\CR} \: \& \: f(\vector{x}^{i,t}) \geq f^{{\rm avg}}_t\\
\CR_{i,t} &   {\rm otherwise}
  \end{cases},
\end{align}
}%
where $\alpha = (f(\vector{x}^{i,t}) - f^{{\rm min}}_t) / (f^{{\rm avg}}_t  - f^{{\rm min}}_t)$, and 
$f^{{\rm avg}}_t$ is the average objective value of individuals in $\vector{P}^t$.
The recommended values of $\tau_F$ and $\tau_{\CR}$ are $0.1$. 

\noindent {\bf $\bullet$ PCM-cDE} (Algorithm S.15):
The PCM of competitive DE (PCM-cDE) \cite{Tvrdik06} adaptively selects a combination of $F$ and $\CR$ values from a pre-defined parameter pool.
Although several variants of PCM-cDE (e.g., \cite{Tvrdik09}) have been proposed, we describe the original version in \cite{Tvrdik06}.
For $F$ and $\CR$, $\vector{F}^{\rm pool} = \{0.5, 0.8, 1\}$ and $\vector{\CR}^{\rm pool} = \{0, 0.5, 1\}$ are defined.
There are nine possible combinations of the $F$ and $\CR$ values from each pool, 
denoted as $\vector{q}^1 = (0.5, 0), ..., \vector{q}^9 = (1, 1)$.


For each $t$, a pair of parameters assigned for each individual $\vector{x}^{i,t}$ is selected from $\vector{q}^1, ..., \vector{q}^9$.
The selection probability $s_{k,t} \in (0,1]$ of selecting $\vector{q}^k$ ($k \in \{1, ..., 9\}$) is given as follows:
{\small
\begin{align}  
  \label{eqn:cde}
s_{k,t} = \frac{n^{\rm succ}_{k} + n^0}{\sum^9_{l=1} (n^{\rm succ}_{l} + n^0)},
\end{align}
}%
 where $n^0$ is a parameter to avoid $s_{k,t}=0$.
In \eqref{eqn:cde}, $n^{\rm succ}_{k}$ represents the number of successful trials of $\vector{q}^{k}$ from the last initialization.
When any $s_{k,t}$ is below the threshold $\delta$, all of the $n^{\rm succ}$ values are reinitialized to $0$.
The recommended settings of $n^0$ and $\delta$ are $2$ and $1/45$, respectively.

\noindent {\bf $\bullet$ PCM-SaDE} (Algorithm S.16):
In the PCM of Self-adaptive DE (PCM-SaDE) \cite{QinHS09}, $F_{i,t}$ and $\CR_{i,t}$ are generated as follows: $F_{i,t} = {\rm randn}(0.5, 0.3)$ and $\CR_{i,t} = {\rm randn}(\mu_{\CR}, 0.1)$.
Even when $F_{i,t}$ is outside of $[0, 1]$,  a repair method is not applied.
%
%
%
The mean $\mu_{\CR}$ is set to $0.5$ at the beginning of the search.
While the values of $F$ are randomly generated, the $\CR$ parameter is adaptively adjusted.
For each iteration, successful $\CR$ parameters are stored into a historical memory $\vector{H}^{\CR}$.
When the number of iterations $t$ exceeds a learning period $t^{{\rm learn}}$, $\mu_{\CR}$ is set to the median value\footnote{In the earlier conference version of PCM-SaDE \cite{QinS05}, the mean value of elements in $\vector{H}^{\CR}$ is used for update of $\mu_{\CR}$.} of all elements in $\vector{H}^{\CR}$.
For example, 
if elements in the historical memory are $\vector{H}^{\CR,1} = \{0.1, 0.2\}$, $\vector{H}^{\CR,2} = \{0.3\}$, and $\vector{H}^{\CR,3} = \{0.4, 0.5, 0.6, 0.7\}$, then 
$\mu_{\CR} $ is set to $0.4$.
When $t > t^{{\rm learn}}$, elements stored at the earliest iteration are removed from the historical memory (i.e., first-in-first-out replacement policy).

\noindent {\bf $\bullet$ PCM-SaNSDE} (Algorithm S.17):
The PCM of Self-adaptive DE with Neighborhood Search (PCM-SaNSDE) \cite{YangTY08} is a hybrid method combining PCM-SaDE (the conference version \cite{QinS05}) and PCM-NSDE \cite{YangYH08}.
While PCM-SaDE randomly generates the values of $F$ according to the Normal distribution, PCM-SaNSDE adaptively selects the probability distribution for generating $F$ as follows:
{\small
\begin{align}  
\label{eqn:sansde-sf}
  F_{i,t} = \begin{cases}
    {\rm randn}(0.5, 0.3) & \:  {\rm if} \: {\rm randu[0,1]} < p\\
    {\rm randc} (0,1) & \:  {\rm otherwise}
  \end{cases},
\end{align}
}%
where the meta-parameter $p \in [0,1]$ controls the probability of selecting the Normal and Cauchy distributions. 
The value of $p$ is initialized to $0.5$.
Then, after the learning period $t^{\rm learn}$, $p$ is updated as follows:
{\small
\begin{align}  
\label{eqn:sansde-fp}
p =  \frac{n^{\rm succ1} (n^{\rm total2})}{n^{\rm succ2} (n^{\rm total1}) + n^{\rm succ1} (n^{\rm total2}) },
\end{align}
}%
where $n^{\rm total1}$ and $n^{\rm total2}$ are the number of times that the Normal and Cauchy distributions are selected during the learning period.
The $n^{\rm succ1}$ and $n^{\rm succ2}$ represent the number of successful trials when using $F$ values generated by each distribution.
Once $p$ is updated, the four parameters ($n^{\rm total1}$, $n^{\rm total2}$, $n^{\rm succ1}$, and $n^{\rm succ2}$) are set to $0$.

$\CR_{i,t}$ is generated according to the procedure of PCM-SaDE, and $\mu_{\CR}$ is updated based on the weighted mean as follows:
{\small
\begin{align}  
  \label{eqn:sansde-cr-up}
\mu_{\CR} = \sum^{|\vector{S}^{\CR}|}_{k=1} w_k S^{\CR}_k, \: w_k = \frac{|f(\vector{x}^k) - f(\vector{u}^k)|}{ \sum^{|\vector{S}^{\CR}|}_{l=1} |f(\vector{x}^l) - f(\vector{u}^l)|}.
%
\end{align}
}%

\noindent {\bf $\bullet$ PCM-JADE} (Algorithm S.18):
The PCM of Jingqiao and Arthur's DE (PCM-JADE) \cite{ZhangS09} uses two adaptive meta-parameters $\mu_{F} \in (0,1]$ and  $\mu_{\CR} \in [0,1]$ for adaptation of $F$ and $\CR$, respectively.
At the beginning of the search, $\mu_{F}$ and $\mu_{\CR}$ are both initialized to $0.5$ and adapted during the search.
For each iteration $t$, $F_{i,t}$ and $\CR_{i,t}$ are generated as follows: $F_{i,t} = {\rm randc}(\mu_{F}, 0.1)$ and $\CR_{i,t} = {\rm randn}(\mu_{\CR}, 0.1)$.
%
When $F_{i,t}> 1$, $F_{i,t}$ is truncated to $1$. When $F_{i,t} \leq 0$, the new $F_{i,t}$ is repeatedly generated  in order to generate a valid value.

For each iteration $t$, successful $F$ and $\CR$ parameters are stored into sets $\vector{S}^{F}$ and $\vector{S}^{\CR}$, respectively. 
We use $\vector{S}$ to refer to $\vector{S}^{F}$ or $\vector{S}^{\CR}$ wherever the ambiguity is irrelevant or resolved by context.
At the end of the iteration, $\mu_{F}$ and $\mu_{\CR}$ are updated as: $\mu_{F} = (1 - c) \: \mu_{F} + c \: {\rm mean}_L(\vector{S}^{F})$ and $  \mu_{\CR} = (1 - c) \: \mu_{\CR} + c \: {\rm mean}_A(\vector{S}^{\CR})$, 
%
where $c \in [0,1]$ is a learning rate, 
${\rm mean}_A(\vector{S})$ is the arithmetic mean of $\vector{S}$, and ${\rm mean}_L(\vector{S})$ is the Lehmer mean of $\vector{S}$ which is computed as: ${\rm mean}_L(\vector{S}) = (\sum_{s \in \vector{S}} s^2) / (\sum_{s \in \vector{S}} s)$.
%



\noindent {\bf $\bullet$ PCM-IMDE} (Algorithm S.19):
The PCM of Islam's Modified DE  (PCM-IMDE) \cite{IslamDGRS12} is similar to PCM-JADE and also uses the meta-parameters $\mu_{F}$ and  $\mu_{\CR}$.
In each iteration $t$, $F_{i,t}$ and $\CR_{i,t}$ are generated as same with PCM-JADE.
At the end of each iteration, $\mu_{F}$ and $\mu_{\CR}$ are updated as follows: $\mu_{F} = (1 - c_F) \: \mu_{F} + c_F \: {\rm mean}_P(\vector{S}^{F})$ and $\mu_{\CR} = (1 - c_{\CR}) \: \mu_{\CR} + c_{\CR} \: {\rm mean}_P(\vector{S}^{\CR})$, 
%
where $c_F$ and $c_{\CR}$ are uniformly selected random real numbers from $[0, 0.2]$ and $[0, 0.1]$, respectively.
Unlike JADE, the learning rates $c_F$ and $c_{\CR}$ are randomly assigned in each iteration $t$.
The function ${\rm mean}_P(\vector{S})$ denotes the power mean of $\vector{S}$: ${\rm mean}_P(\vector{S}) = \bigr( \frac{1}{|\vector{S}|} \sum_{s \in \vector{S}} s^{1.5} \bigl)^{(1/1.5)}$.
%
%
%

\noindent {\bf $\bullet$ PCM-SHADE} (Algorithm S.20):
Similar to PCM-SaDE, the PCM of Success-History based Adaptive DE (PCM-SHADE) \cite{TanabeF13,TanabeF14CEC} uses historical memories  $\vector{M}^{F} $ and $ \vector{M}^{\CR}$ for adaption of $F$ and $\CR$, where $\vector{M}^{F} = (M^{F}_1, ..., M^{F}_H)$ and $\vector{M}^{\CR} = (M^{\CR}_1, ..., M^{\CR}_H)$.
Here, $H$ is a memory size, and all elements in $\vector{M}^{F} $ and $ \vector{M}^{\CR}$ are initialized to $0.5$.
Although there have been several slightly different variants of PCM-SHADE \cite{TanabeF13,TanabeF14CEC,TanabeF17}, the simplest version in \cite{TanabeF17} is described here.
In each iteration $t$,  $F_{i,t}$ and $\CR_{i,t}$ are generated as follows: $F_{i,t} = {\rm randc}(M^{F}_{r_{i,t}}, 0.1)$ and $\CR_{i,t} = {\rm randn}(M^{\CR}_{r_{i,t}}, 0.1)$, where $r_{i,t}$ is a randomly selected index from $\{1, ..., H\}$.
%
%
%
%
If $F_{i,t}$ and $\CR_{i,t}$ are outside the range $[0, 1]$, they are adjusted/regenerated according to the procedure of  PCM-JADE.

At the end of each iteration, the memory contents in $\vector{M}^{F}$ and $\vector{M}^{\CR}$ are updated using the Lehmer mean as follows: $M^{F}_k = {\rm mean}_L(\vector{S}^{F})$ and $M^{\CR}_k = {\rm mean}_L(\vector{S}^{\CR})$.
%
%
%
An index $k \in \{1, ..., H\}$ determines the position in the memory to update.
At the beginning of the search, $k$ is initialized to $1$, 
and incremented whenever a new element is inserted into the history.
If $k > H$, $k$ is set to $1$.


\noindent {\bf $\bullet$ PCM-SLADE} (Algorithm S.21):
The PCM of SLADE (PCM-SLADE) \cite{ZhaoYHC16} is similar to PCM-JADE.
However, contrary to PCM-JADE, the values of $F$ and $\CR$ are randomly generated according to the Normal and Cauchy distributions as follows: $F_{i,t} = {\rm randn}(\mu_{F}, 0.1)$ and $ \CR_{i,t} = {\rm randc}(\mu_{\CR}, 0.1)$.
%
%
%
When $F_{i,t}$ is outside of $[0,1]$, it is set to $1$.
The $\CR_{i,t}$ is repeatedly generated until it is inside of $[0,1]$.
The arithmetic mean is used for updating both $\mu_{F}$ and $\mu_{\CR}$ as follows: $\mu_{F} = (1 - c) \: \mu_{F} + c \: {\rm mean}_A(\vector{S}^{F})$ and $  \mu_{\CR} = (1 - c) \: \mu_{\CR} + c \: {\rm mean}_A(\vector{S}^{\CR})$.
This differs slightly from the original description in \cite{ZhaoYHC16}.
For details, see Section S.2 in the supplementary file.


\noindent {\bf $\bullet$ PCM-EPSDE} (Algorithm S.22):
The PCM of EPSDE (PCM-EPSDE) \cite{MallipeddiSPT11} uses an $\vector{F}^{\rm pool}$ and a $\vector{\CR}^{\rm pool}$ for  adaptation of $F$ and $\CR$,  respectively.
The $\vector{F}^{\rm pool}$ and $\vector{\CR}^{\rm pool}$ are sets of the $F$ and $\CR$ values as follows: $\vector{F}^{\rm pool} = \{0.4, 0.5, ..., 0.9\}$ and $\vector{\CR}^{\rm pool} = \{0.1, 0.2, ..., 0.9\}$.
At the beginning of the search, each individual $\vector{x}_{i,t}$ is randomly assigned values for $F_{i,t}$ and $\CR_{i,t}$ from each pool.
During the search, successful parameter sets are inherited by the individual in the next iteration $t+1$.
Parameter sets that fail are reinitialized.

\noindent {\bf $\bullet$ PCM-CoBiDE} (Algorithm S.23):
In the PCM of DE with covariance matrix learning and bimodal distribution parameter setting (PCM-CoBiDE) \cite{WangLHL14}, at the beginning of the search, $F_{i,t}$ and $\CR_{i,t}$ are randomly sampled according to a bimodal distribution consisting of two Cauchy distributions as follows:
{\small
\begin{align}
\label{eqn:bimodal_f}
  F_{i,t} &= \begin{cases}
    {\rm randc} (0.65,0.1) & \:  {\rm if} \: {\rm randu[0,1]} < 0.5\\
    {\rm randc} (1.0,0.1) & \:  {\rm otherwise}
  \end{cases},
\\
\label{eqn:bimodal_cr}
  \CR_{i,t} &= \begin{cases}
    {\rm randc} (0.1,0.1) & \:  {\rm if} \: {\rm randu[0,1]} < 0.5\\
    {\rm randc} (0.95,0.1) & \:  {\rm otherwise}
  \end{cases},
\end{align}
}%
where $F_{i,t}$ and $\CR_{i,t}$ are outside of $[0,1]$, they are modified according to the procedure of PCM-JADE.
Similar to PCM-EPSDE, a pair of successful $F$ and $\CR$ values is inherited by each individual in the next iteration. 
Failed $F$ and $\CR$ parameter pairs are reinitialized using \eqref{eqn:bimodal_f} and \eqref{eqn:bimodal_cr}.


\noindent {\bf $\bullet$ PCM-DEDPS} (Algorithm S.24):
The PCM of DE with Dynamic Parameters Selection (PCM-DEDPS) \cite{SarkerER14} uses pre-defined parameter combinations of $F$ and $\CR$ similar to PCM-cDE, but the pool of parameter combinations is pruned during the search process.
For $F$ and $\CR$, the following sets of candidate parameters are prepared at the beginning of the search: $\vector{F}^{\rm pool} = \{0.4, 0.5, ... , 0.9, 0.99\}$, $\vector{\CR}^{\rm pool} = \{0.2, 0.3, ..., 0.9, 0.99\}$.
The number of possible combination of the $F$ and $\CR$ values $m$ is $7 \times 9 = 63$.
At the beginning of each iteration, all possible pairs $\vector{q}^1, ..., \vector{q}^{m}$ are randomly assigned to all individuals without replacement.
If $N >m$, randomly selected pairs are assigned to the $N-m$ individuals.

When $t$ reaches a learning period $t^{\rm CS}  \in \{50, 100, 150, 200\}$, each pair $\vector{q}^{k}$ ($k \in \{1, ..., m\}$) is ranked based on its score value $n^{\rm score}_{k}$ calculated as follows: $n^{\rm score}_{k}  = n^{\rm succ}_{k} / n^{\rm total}_{k}$.
%
%
%
%
%
The $n^{\rm total}_{k}$ is the number of times that $\vector{q}^k$ was selected, and $n^{\rm succ}_{k}$ denotes the number of successful trials of $\vector{q}^k$.
A high $n^{\rm score}_{k}$ indicates that $\vector{q}^k$ is an appropriate setting.
All pairs $\vector{q}^1, ..., \vector{q}^{m}$ are sorted based on their score values, and then their lower half is removed from the pool of possible parameter combinations.
After the pruning procedure, the two parameters ($n^{\rm succ}_{k}$ and $n^{\rm total}_{k}$) for each $\vector{q}^k$ are reinitialized to $0$.


\noindent {\bf $\bullet$ PCM-RDE} (Algorithm S.25):
In the PCM of Rank-based DE (PCM-RDE) \cite{TakahamaS12a}, for each iteration $t$, individuals are sorted based on their objective values so that $f(\vector{x}^{1,t}) \leq ... \leq f(\vector{x}^{N,t})$.
Then, $F_{i,t}$ and $\CR_{i,t}$ are assigned according to the rank value of the base vector $j \in \{1, ..., N\}$:
{\small
\begin{align}
\label{eqn:rde_f}
\small
F_{i,t}  &= F^{\rm min} + (F^{\rm max}  - F^{\rm min}) \biggl(\frac{j - 1}{N-1}\biggr),\\
\label{eqn:rde_cr}
\CR_{i,t}  &= \CR^{\rm max} - (\CR^{\rm max}  - \CR^{\rm min}) \biggl(\frac{j - 1}{N-1}\biggr),
\end{align}
}%
where $F^{\rm min}$, $F^{\rm max}$, $\CR^{\rm min}$, and $\CR^{\rm max}$ are the minimum and maximum values for $F$ and $\CR$, respectively.
Their recommended settings are as follows: $F^{\rm min} = 0.6$, $F^{\rm max} = 0.95$, $\CR^{\rm min} = 0.85$, and $\CR^{\rm max} = 0.95$.
In \eqref{eqn:rde_f} and \eqref{eqn:rde_cr}, a smaller $F_{i,t}$ value and a larger $\CR_{i,t}$ value are assigned when the objective value of the base vector is small.
While other PCMs described in this section were originally developed for a DE algorithm using binomial crossover, PCM-RDE was designed for a DE with exponential crossover.

\noindent {\bf $\bullet$ PCM-IDE} (Algorithm S.26):
The PCM of DE with an individual dependent mechanism (PCM-IDE) \cite{TangDL15} uses the rank values of individuals in the population similar to PCM-RDE.
However, while values of $F$ and $\CR$ are deterministically assigned to individuals based on their ranks in PCM-RDE, they are randomly generated in PCM-IDE.
After sorting all individuals based on their objective values, $F_{i,t}$ and $\CR_{i,t}$ values for an individual $\vector{x}^{i,t}$ are sampled as follows: $F_{i,t}  = {\rm randn}(\mu_{F,j}, 0.1)$ and $\CR_{i,t}  = {\rm randn}(\mu_{\CR,i}, 0.1)$, 
%
%
where $\mu_{F,j} = j/N$ ($j \in \{1, ..., N\}$) is the rank value of a base vector $\vector{x}^{j,t}$, 
and $\mu_{\CR,i}$ ($i \in \{1, ..., N\}$) for the $i$-th ranked individual is $i/N$.
The $F_{i,t}$ and $\CR_{i,t}$ values are repeatedly generated until they are in  $[0, 1]$.
In contrast to PCM-RDE, a small $\CR_{i,t}$ value is assigned to better individuals in the population.

\noindent {\bf $\bullet$ PCM-YADE} (Algorithm S.27):
In the PCM of Yu's Adaptive DE (PCM-YADE) \cite{YuSCZGLLZ14}, for each iteration $t$, the current search state is classified into the exploration and exploitation phase based on the distribution of individuals in the objective and solution spaces.
Then, $F_{i,t}$ and $\CR_{i,t}$ are randomly generated according to the current state.
The description of the PCM-YADE procedure requires numerous equations -- due to space, we show the details in the supplement (Section S.1).

\subsection{SPCMs in DE}
\label{sec:spcm}

\noindent {\bf $\bullet$ PCM-SDE} (Algorithm S.28):
In the PCM of Self-adaptive DE (PCM-SDE) \cite{OmranSE05}, $\CR_{i,t}$ is randomly sampled as follows: $\CR_{i,t} = {\rm randn}(0.5, 0.15)$.
Unlike $\CR$, $F_{i,t}$ is self-adaptively generated as follows: $F_{i,t} = F_{r_1,t} + {\rm randn}(0, 0.5) (F_{r_2,t} - F_{r_3,t})$.
%
At the beginning of the search, $F_{i,t}$ is initialized according to a Normal distribution  ${\rm randn}(0.5, 0.15)$.
The indices $r_1$, $r_2$, and $r_3$ are randomly selected from $\{1, ..., N\}$ such that they differ from each other.
Clearly, the generation method of new $F$ values is identical to the rand/1 mutation strategy.
Values of $F_{i,t}$ and $\CR_{i,t}$  outside of $[0,1]$ are truncated, e.g., 
$F_{i,t} = 1.4$ is truncated to $0.4$. 


\section{Relationships Among PCMs in DE}
\label{sec:discussion-pcms}


We discuss the relationships among PCMs for DE, including the 24 PCMs described in Section \ref{sec:pcm}.
%
First, we provide a historical overview of the early development of PCMs in Section \ref{sec:discussion_early_studies}.
Then, we provide more detailed classifications of PCMs in Sections \ref{sec:discussion_classification}.
DPCMs and SPCMs are described in Sections \ref{sec:discussion_dpcms} and \ref{sec:discussion_spcms}, respectively.

Then, we turn to discussions of APCMs, which are currently the most common class of PCMs and the main focus of this section. 
First, observation-based APCMs are described in Section \ref{sec:discussion_observation-based_apcms}.
Then, success-based APCMs are discussed in the remaining sections.
APCMs with a parameter inheritance mechanism are described in Section \ref{sec:discussion_inheritance}.
APCMs with predefined parameters sets are introduced in Section \ref{sec:discussion_pcm_predefined}.
PCM-SaDE and PCM-JADE variants are introduced in Sections \ref{sec:discussion_pcmsade} and \ref{sec:discussion_pcmjade}, respectively.
Finally, Section \ref{sec:discussion_pcmshade} discusses PCM-SHADE, which is used in many state-of-the-art DEs and integrates key ideas from the two previously independent lines of work described in Sections \ref{sec:discussion_pcmsade} and \ref{sec:discussion_pcmjade}.


\subsubsection{Historical perspective}
\label{sec:discussion_early_studies}

Most of PCMs in DE proposed in early studies are derived from those in other EAs such as GA and PSO.
For example, PCM-DERSF, PCM-DETVSF, as well as some other methods (e.g., \cite{KaeloA06}) are based on PCMs of two PSO variants (random and time-varying inertia weight strategies) \cite{Shi99,EberhartS01}.
The idea of FADE \cite{LiuL05}, which adaptively adjusts the $F$ and $\CR$ parameters using fuzzy logic controllers, is derived from Fuzzy GAs \cite{LeeT93}.
DESAP \cite{Teo06}, which adapts the three control parameters ($F$, $\CR$, and the population size $N$), is also based on self-adaptive GAs.

%

A significant turning point in the development of PCMs for DE was around 2005.
PCM-SaDE \cite{QinS05} (the conference version) and PCM-jDE \cite{BrestGBMZ06} were proposed in 2005 and 2006, respectively.
The two PCMs were designed based on the unique algorithmic characteristics of DE (i.e., the pair-wise survival selection of individuals for the next iteration). 
PCM-jDE assigns a pair of $F_{i,t}$ and $\CR_{i,t}$ to each individual $\vector{x}^{i,t}$, and each parameter is randomly regenerated with a predefined probability as \eqref{eqn:jde-f} and \eqref{eqn:jde-cr}.
If the trial vector $\vector{u}^{i,t}$ is better than its parent individual $\vector{x}^{i,t}$, the newly generated parameter is inherited by $\vector{x}^{i,t+1}$.
PCM-SaDE samples $\CR_{i,t}$ for each individual $\vector{x}^{i,t}$ according to the Normal distribution with the mean $\mu_{\CR}$ and variance $0.1$.
The parameter 
$\mu_{\CR}$ is adaptively updated based on successful $\CR$ values in the historical memory $\vector{H}^{\CR}$.
Recall that the determination of success or failure is made according to a pair-wise comparison between $\vector{x}^{i,t}$ and $\vector{u}^{i,t}$.
Such APCMs based on the comparison between two individuals had not been proposed for {\em population-based} EAs\footnote{The step-size adaptation of one-fifth success rule for $(1+1)$-ES \cite{Rechenberg73} is also based on the pair-wise comparison between a parent individual $\vector{x}^{t}$ and a child $\vector{u}^{t}$. However, $(1+1)$-ES uses only one individual for the search.}.

\subsubsection{A more detailed classification of PCMs in DE}
\label{sec:discussion_classification}

Table \ref{tab:pcms} shows the properties of the 24 PCMs.
Each symbol in Table \ref{tab:pcms} is explained below.
Although Table \ref{tab:pcms} attempts to clearly organize the 24 PCMs reviewed in this paper,
we do not claim that this categorization can be applied to all PCMs.
A systematic taxonomy which can be used to comprehensively classify the myriad of PCMs for DE in the literature remains an open problem.
In Section \ref{sec:pcm}, PCMs in DE are classified into DPCMs, APCMs, and SPCMs according to the taxonomy of Eiben et al. \cite{EibenHM99}.
Although the classification rule in \cite{EibenHM99} has been widely accepted in the evolutionary computation community, some previous studies introduce other taxonomies that categorize PCMs for DE in detail.


According to \cite{TakahamaS12a}, APCMs in DE can be further classified into the following two categories: ``S'' success-based control and ``O'' observation-based control.
While success-based APCMs adjust parameter values of $F$ and $\CR$ based on the success/failure decision, observation-based APCMs adapt parameter values using some indicator values of individuals in the solution and/or objective spaces.

%

Another taxonomy in \cite{ChiangCL13} categorizes PCMs based on the following four factors: the type of parameter values (``C'' continuous or ``D'' discrete values), the number of parameter values for each iteration (``S'' a single value or ``M'' multiple values), the information used while sampling new parameter values, and the parameter inheritance mechanism.
For the third factor, the following four information resources can be considered: ``N'' no information, ``T'' time, ``P'' the distribution of individuals in the population, and ``H'' the historical information obtained during the search process.
Note that this description of their classification differs slightly from the original one in \cite{ChiangCL13}.
Since some factors in \cite{ChiangCL13} can be represented by the taxonomy in \cite{EibenHM99}, we modified the original taxonomy for the sake of clarity.


As seen from Table \ref{tab:pcms}, while most APCMs can be categorized as either a success-based or an observation-based control method, PCM-FDSADE and PCM-ISADE are exceptional and belong to both categories.
Although PCM-FDSADE and PCM-ISADE sample new $F$ and $\CR$ values based on the objective values of individuals, they use the success/failure determination to select $F$ and $\CR$ values which are inherited to the next iteration.
Five PCMs (PCM-CoDE, PCM-SWDE, PCM-cDE, PCM-EPSDE, and PCM-DEDPS) use predefined discrete parameter values.
While a single parameter value is commonly used in all the individuals in three PCMs (PCM-DETVSF, PCM-SinDE, and PCM-DEPD), the remaining 21 PCMs use different values of $F$ and $\CR$ when producing offspring in an iteration. 
Two DPCMs (PCM-DETVSF and PCM-SinDE) are time-dependent PCMs, and other PCMs utilize various types of information to sample new parameter values.
In five APCMs (three PCM-jDE and two PCM-EPSDE variants) and one SPCM (PCM-SDE), the $F$ and $\CR$ values used in each individual can be inherited by the next iteration.






\begin{table}
\begin{center}
  \caption{\small The properties of 24 PCMs for DE. Each symbol is explained in Section \ref{sec:discussion_classification}.}
{\footnotesize
  \label{tab:pcms}
\scalebox{0.89}[1]{ 
\begin{tabular}{lccccccccc}
\midrule
& $F$ & $\CR$ & S/O & C/D & S/M & Info. & Inh.\\
\toprule
%
PCM-DERSF \cite{DasKC05a} & DPCM & 0.9 & & C & M & N & \\
PCM-DETVSF \cite{DasKC05a} & DPCM & 0.9 & & C & S & T & \\
PCM-CoDE \cite{WangCZ11} & DPCM & DPCM & & D & M & N & \\
PCM-ZMDE \cite{ZouWGL13} & DPCM & DPCM & & C & M & N & \\
PCM-SinDE \cite{DraaBB15} & DPCM & DPCM & & C & S & T & \\
PCM-SWDE \cite{DasGM15} & DPCM & DPCM & & D & M & N & \\
\midrule
PCM-DEPD \cite{AliT04} & APCM & 0.5 & O & C & S & P & \\
PCM-jDE \cite{BrestGBMZ06} & APCM & APCM & S & C & M & N & $\checkmark$\\
PCM-cDE \cite{Tvrdik06} & APCM & APCM & S & D & M & H & \\
PCM-SaNSDE \cite{YangTY08} & APCM & APCM & S & C & M & H & \\
PCM-FDSADE \cite{TirronenN09} & APCM & APCM & S\&O & C & M & P & $\checkmark$\\
PCM-ISADE \cite{JiaGW09} & APCM & APCM & S\&O & C & M & P & $\checkmark$\\
PCM-SaDE \cite{QinHS09} & DPCM & APCM & S & C & M & H & \\
PCM-JADE \cite{ZhangS09} & APCM & APCM & S & C & M & H & \\
PCM-EPSDE \cite{MallipeddiSPT11} & APCM & APCM & S & D & M & N &$\checkmark$ \\
PCM-RDE \cite{TakahamaS12a} & APCM & APCM & O & C & M & P & \\
PCM-IMDE \cite{IslamDGRS12} & APCM & APCM & S & C & M & H & \\
PCM-SHADE \cite{TanabeF13} & APCM & APCM & S & C & M & H & \\
PCM-YADE \cite{YuSCZGLLZ14} & APCM & APCM & O & C & M & P & \\
PCM-DEDPS \cite{SarkerER14} & APCM & APCM & S & D & M & H & \\
PCM-CoBiDE \cite{WangLHL14} & APCM & APCM & S & C & M & N & $\checkmark$\\
PCM-IDE \cite{TangDL15} & APCM & APCM & O & C & M & P & \\
PCM-SLADE \cite{ZhaoYHC16} & APCM & APCM & S & C & M & H & \\
\midrule
PCM-SDE \cite{OmranSE05} & SPCM & 0.5 & S & C & M & P  & $\checkmark$\\
\midrule
\end{tabular}
}
}
 \end{center}
\end{table}

\subsubsection{DPCMs}
\label{sec:discussion_dpcms}

PCM-DERSF and PCM-DETVSF, which are classified as DPCMs, were proposed in 2005.
However, DPCMs did not become popular 
in the DE community until PCM-CoDE was proposed in 2011.
``CoDE'' is a simple but efficient DE algorithm with a DPCM, and experimental results in \cite{WangCZ11} show that it performs better than four DE algorithms with APCMs.
After the proposal of PCM-CoDE, some efficient DPCMs (e.g., PCM-ZMDE, PCM-SinDE, and PCM-SWDE) were proposed.
While PCM-DERSF and PCM-DETVSF adjust only values of $F$, four DPCMs (PCM-CoDE, PCM-ZMDE, PCM-SinDE, and PCM-SWDE) control both $F$ and $\CR$ parameters.
Moreover, PCM-CoDE, PCM-SinDE, and PCM-SWDE generate values of $F$ and $\CR$ in a wider range than PCM-DERSF and PCM-DETVSF.
PCM-CoDE and PCM-SWDE are similar in that they randomly assign a predefined parameter pair $\{F, \CR\}$ to each individual (see Table \ref{tab:pcms}).
%
It is interesting to note that an adaptive version of PCM-CoDE based on the PCM-SaDE scheme is proposed in \cite{WangCZ11}.
However, it performs significantly worse than the simple, original PCM-CoDE in the same DE framework.

\subsubsection{SPCMs}
\label{sec:discussion_spcms}


SPCMs have not received much attention in the DE community, so
only one SPCM (PCM-SDE) is described in Section \ref{sec:spcm}.
In addition to PCM-SDE, PCMs in ``SPDE'' \cite{Abbass02}, ``DESAP'' \cite{Teo06}, ``ESADE'' \cite{GuoLLSWC14}, and ``PBMODE'' \cite{GuoY18} can be classified into SPCMs\footnote{A method of adjusting a weight factor used in a mutation strategy in DEGL \cite{DasACK09} belongs to SPCMs.}.
However, these algorithms do not fit well within the general DE framework covered in this paper.
Unlike a standard DE, GA mutation operators are incorporated into SPDE and DESAP.
The PCM of ESADE needs to generate two trial vectors for each individual.
A Pareto-dominance relation is used in the PCM of PBMODE.
Thus, it is difficult to extract only the PCM from SPDE, DESAP, ESADE, and PBMODE.


\subsubsection{Observation-based APCMs}
\label{sec:discussion_observation-based_apcms}


Observation-based APCMs use indicator values from the individuals and/or their objective values for adaptation of $F$ and $\CR$.
In addition to the four APCMs in Table \ref{tab:pcms}, other observation-based APCMs have been proposed, including PCM-FADE \cite{LiuL05} and PCM-FiADE \cite{GhoshDCG11}.


Interestingly, each observation-based APCM was designed based on different search policies.
While PCM-DEPD, PCM-FiADE, PCM-RDE, and PCM-IDE use only information from the objective space, PCM-FADE and PCM-YADE assign values of $F$ and $\CR$ to individuals based on their diversity in both the objective and solution spaces.
In PCM-RDE and PCM-FiADE, small $\CR$ values are assigned to each individual in case of a selected base vector is {\em inferior} in the population.
In contrast, PCM-IDE assigns small $\CR$ values to {\em superior} individuals.
Thus, the key to designing an efficient observation-based APCM is how the information obtained from the distribution of individuals in the population is used.




\subsubsection{APCMs with a parameter inheritance mechanism}
\label{sec:discussion_inheritance}


Some variants of PCM-jDE have been proposed, such as PCM-FDSADE and PCM-ISADE, as well as other methods (e.g., \cite{NomanBI11}).
Although $F_{i,t}$ and $\CR_{i,t}$ for $\vector{x}_{i,t}$ are probabilistically changed to new values in PCM-jDE variants, this probability and the method of generating values of $F$ and $\CR$ are different for each PCM, as shown in \eqref{eqn:jde-f}--\eqref{eqn:isade-cr}.
While the probability is constant in PCM-jDE, it depends on the diversity of the objective values of individuals in PCM-FDSADE.
PCM-ISADE utilizes the objective value of each individual to determine the probability and generate new values of $F$ and $\CR$.





Since the parameter assignment strategy of PCM-CoBiDE is identical with that of PCM-EPSDE, PCM-CoBiDE is considered to be a variant of PCM-EPSDE.
While PCM-EPSDE uses a predefined parameter pool, PCM-CoBiDE generates parameters with a bimodal distribution as \eqref{eqn:bimodal_f} and \eqref{eqn:bimodal_cr}.
In addition to PCM-CoBiDE, some PCM-EPSDE variants have been proposed (e.g., \cite{IaccaCN15}).

Although PCM-jDE and PCM-EPSDE are similar in that they assign parameter values of $F$ and $\CR$ to each individual, 
their adaptation strategies differ.
While PCM-jDE uses new parameters probabilistically, PCM-EPSDE assigns new parameter values when a trial vector generation fails.
An analysis in \cite{TanabeF17} shows that there is a significant difference in the parameter adaptation ability of PCM-jDE and PCM-EPSDE.





\subsubsection{APCMs using predefined discrete parameters sets}
\label{sec:discussion_pcm_predefined}


In three APCMs (PCM-cDE, PCM-EPSDE, and PCM-DEDPS), discrete values of $F$ and $\CR$ are defined before the search and adaptively assigned to each individual in the population.
However, their parameter adaptation mechanisms are significantly different from each other.
PCM-cDE assigns each pair of parameter values to individuals with a probability based on the number of its successful trials as \eqref{eqn:cde}.
PCM-EPSDE randomly assigns a pair of $F$ and $\CR$ values to an individual at the beginning of the search. The parameter values for each individual are reassigned when the trial is a failure.
On each iteration, PCM-DEDPS assigns all the combinations of $F$ and $\CR$ values uniformly to all the individuals in the population.
Then, parameter combinations which perform poorly regarding the number of successful trials are removed periodically.


Since this kind of APCMs relies entirely on the predefined set of discrete parameters for adaptation, their performance is influenced by the elements of these predefined parameters sets.
The effect of the parameters set in PCM-DEDPS is investigated in \cite{SarkerER14}.
PCM-cDE and PCM-EPSDE variants with other parameter sets have been proposed (e.g., \cite{Tvrdik09,MallipeddiS10SEMCCO}).



\subsubsection{PCM-SaDE variants}
\label{sec:discussion_pcmsade}

As mentioned in Section \ref{sec:discussion_early_studies}, PCM-SaDE is one of the earliest APCMs in DE.
Nevertheless, only a few variants of PCM-SaDE (i.e., PCM-SaNSDE and PCM-SHADE) have been proposed.
This may be because the mutation strategy adaptation in ``SaDE'' attracted more attention than its parameter adaptation mechanism.

PCM-SaNSDE \cite{YangTY08} is an extended version of the conference version of PCM-SaDE \cite{QinS05}.
While PCM-SaDE randomly generates the values of $F$ according to the Normal distribution, PCM-SaNSDE adaptively selects the probability distribution for generating $F$.
Interestingly, it is pointed out in \cite{Dick10} that the adaptive scheme of PCM-SaNSDE for $F$ in \eqref{eqn:sansde-sf} can be replaced with a simple deterministic generation method on some problems without performance loss.

Similar to PCM-SaDE, PCM-SHADE uses the historical memory for adaptation of $F$ and $\CR$.
However, the memory update rule and the method for generating $F$ and $\CR$ values in PCM-SHADE are different from that of PCM-SaDE.


\subsubsection{PCM-JADE variants}
\label{sec:discussion_pcmjade}


PCM-JADE has had a significant impact on the development of PCMs in DE.
%
In addition to the three APCMs (PCM-IMDE, PCM-SHADE, and PCM-SLADE) reviewed in Section \ref{sec:apcm}, a number of PCM-JADE variants have been proposed (e.g., PCM-GaDE \cite{YangTY11}, PCM-MADE \cite{ChengZCN15}, PCM-DMPSADE \cite{FanY15}, and PCM-ZEPDE \cite{FanY16}).

PCM-JADE variants use meta-parameters such as $\mu_{F}$ and $\mu_{\CR}$ as mean/location parameter values of random distributions to generate $F$ and $\CR$ values.
The meta-parameters are updated based on successful $F$ and $\CR$ values at the end of each iteration.
While PCM-JADE uses a constant learning rate $c$, PCM-IMDE randomly generates $c_{F}$ and $c_{\CR}$ values for $F$ and $\CR$ in predefined intervals, respectively.
Unlike PCM-JADE, PCM-SLADE uses the Normal and Cauchy distributions for generating the values of $F$ and $\CR$, respectively.


\subsubsection{PCM-SHADE in state-of-the-art DEs}
\label{sec:discussion_pcmshade}

PCM-SHADE combines two key ideas from the lines of work described above: 
the use of a success-history memory from PCM-SaDE (Section \ref{sec:discussion_pcmsade}), and 
the idea from PCM-JADE of generating new values for $F$ and $C$ by sampling regions centered around previously good values  (Section \ref{sec:discussion_pcmjade}).
The successful combination of these previously disparate lines of work 
led to a significant improvement in the performance of state-of-the-art DE algorithms.
Since 2014, DEs with PCM-SHADE (e.g., L-SHADE \cite{TanabeF14CEC}) have dominated the annual IEEE CEC competitions on single-objective continuous optimization  (\url{http://www3.ntu.edu.sg/home/EPNSugan/index_files/}).
A number of high-performance DE algorithms with PCM-SHADE have been proposed in the past several years \cite{PiotrowskiN18}.

\section{Benchmarking PCMs}
\label{sec:experimental-results}

In this section, we investigate the performance of the 24 PCMs described in Section \ref{sec:pcm}.
In general, it is difficult to directly evaluate the parameter control ability of PCMs. 
Therefore, we evaluate the search performance of DE algorithms using various operators and the 24 PCMs, and treat these results as a proxy for the performance of the 24 PCMs. 
To implement these simplified DE algorithms, we started with the basic, baseline DE \cite{StornP97} described in Section \ref{sec:de} and made the minimal modifications to accommodate each PCM.
For the detail of each DE, see Algorithms S.5--S.28 in the supplementary file.
Below, we refer to ``the search performance of a complex DE with a PCM'' as ``the performance of a PCM'' for simplicity.
Also, ``a complex DE with a PCM'' and ``a PCM'' are used synonymously in this section.

Section \ref{sec:exp-setting} describes the experimental settings.
Section \ref{sec:results-pcm-de-default} compares the performance of PCMs using hyperparameter settings recommended in the literature (usually the original paper), while 
Section \ref{sec:results-pcm-de-tuned} compares the performance of PCMs using hyperparameter settings obtained using SMAC \cite{HutterHL11}. 
Section \ref{sec:results-pcm-vs-gao} investigates the  gap between the performance of existing PCMs with that of a lower bound on the performance of an ``optimal'' PCM obtained by GAO \cite{TanabeF16}.

\subsection{Experimental settings}
\label{sec:exp-setting}

All experiments were conducted using the COCO software (\url{http://coco.gforge.inria.fr/}), the standard benchmarking platform used in the Black Box Optimization Benchmarking (BBOB) workshops held at GECCO (2009-present) and CEC (2015).
We used the noiseless BBOB benchmark set \cite{hansen2012fun}, consisting of 24 test functions $f_1, ...,  f_{24}$.
The 24 BBOB functions are grouped into 5 categories: separable functions ($f_1, ...,  f_5$),  functions with low or moderate conditioning ($f_6, ..., f_9$),  functions with high conditioning and unimodal ($f_{10}, ..., f_{14}$), multimodal functions with adequate global structure ($f_{15}, ..., f_{19}$), and multimodal functions with weak global structure ($f_{20}, ...,  f_{24}$).
The dimensionality $D$ of the functions was set to $2, 3, 5, 10, 20$, and $40$.
For each problem instance, 15 runs were performed.
These settings adhere to the standard benchmarking and analysis procedure adopted by the BBOB community in the BBOB workshops since 2009.
The maximum number of function evaluations was $10\,000 \times D$.

In addition to the 24 PCMs, we evaluated the performance of the DE without any PCMs \cite{StornP97}.
The $F$ and $\CR$ parameters were set to $0.5$ and $0.9$, respectively.
These parameter settings are commonly used in previous work (e.g., \cite{BrestGBMZ06,ZhangS09}).
Below, the DE with $F=0.5$ and $\CR=0.9$ is denoted as ``$F05\CR09$''. 


Following previous work \cite{PosikK12a}, the population size $N$ was set to $5 \times D$ for  $D\geq 5$, and 20 for  $D\leq 3$.
We evaluated the performance of the 25 methods (the 24 PCMs and $F05\CR09$) using eight different mutation operators: rand/1, rand/2, best/1, best/2, current-to-rand/1, current-to-best/1, current-to-$p$best/1, and rand-to-$p$best/1.
Details of the mutation operators are in Table S.1 in the supplementary file.
The control parameters of the current-to-$p$best/1 and rand-to-$p$best/1 strategies were set to $p = 0.05$ and $|\vector{A}| = N$ as in \cite{ZhangS09}.
We evaluated both binomial crossover (bin) and shuffled exponential crossover (sec).
In summary, our benchmarking study investigated the performance of the 25 methods using 16 variation operators (8 mutation strategies $\times$ 2 crossover methods).


Restarts are standard practice in the BBOB community and the COCO software documentation recommends the use of restarts, so  
we incorporated the restart strategy of \cite{TanabeF15} (a slightly modified version of the method in \cite{ZhabitskyZ13}) into all methods, except for PCM-SinDE and PCM-DETVSF.\footnote{In our preliminary experiments, PCM-SinDE and PCM-DETVSF without the restart strategy performed better than with the restart strategy.
PCM-SinDE and PCM-DETVSF gradually adjust control parameters according to the number of iterations, but the restart strategy disturbs their control schedules by reinitializing the population. 
For this reason, time-dependent PCMs are incompatible with the restart strategy.} 
Section S.3 in the supplement describes the restart strategy used.






\begin{figure}[t]
\newcommand{\widthvar}{0.5}
\newcommand{\hspacevar}{-1.5em}
  \begin{center} 
\subfloat[rand/1/bin]{
\includegraphics[width=\widthvar\textwidth]{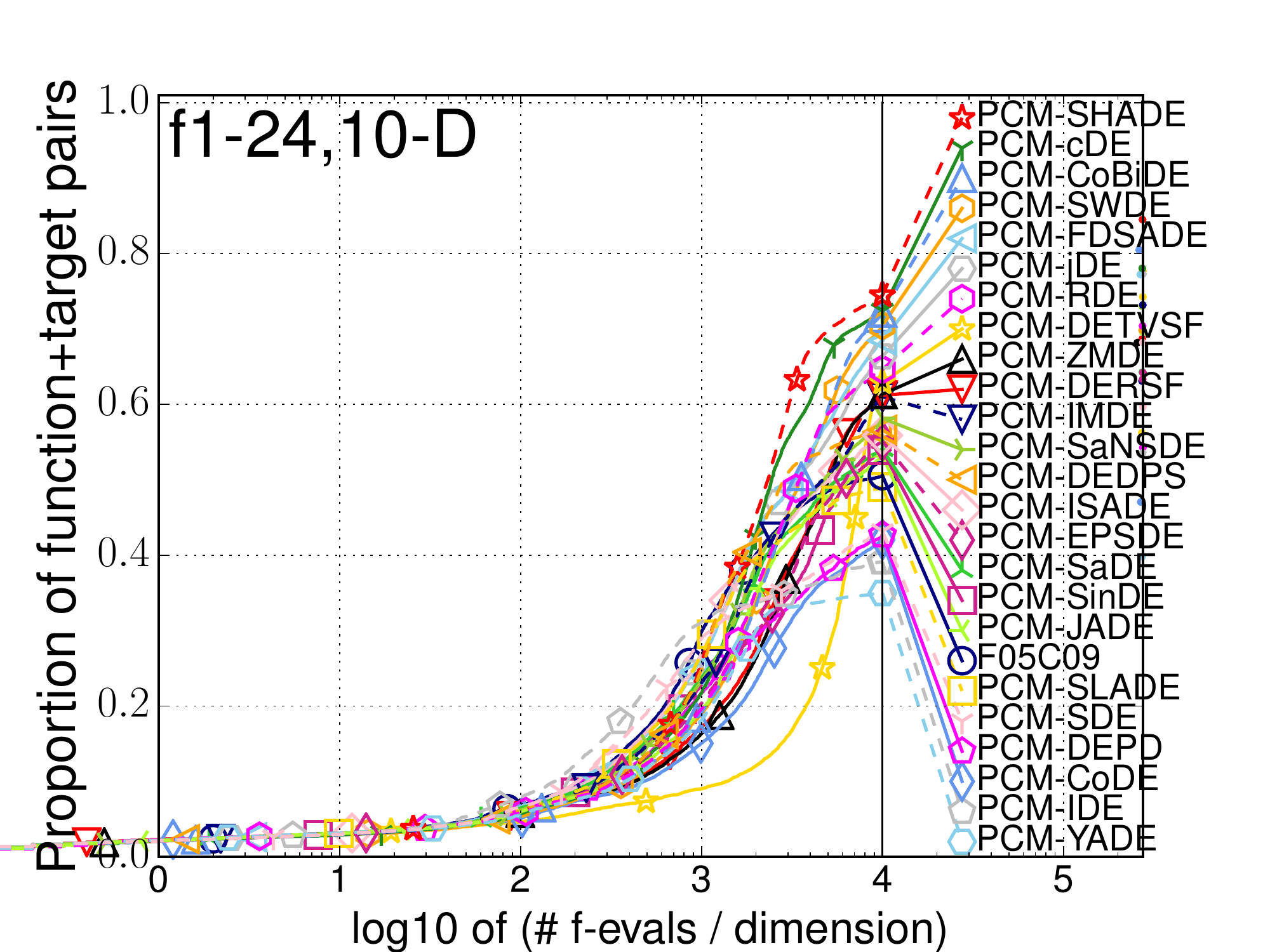}
}
\\
\subfloat[current-to-$p$best/1/bin]{
\includegraphics[width=\widthvar\textwidth]{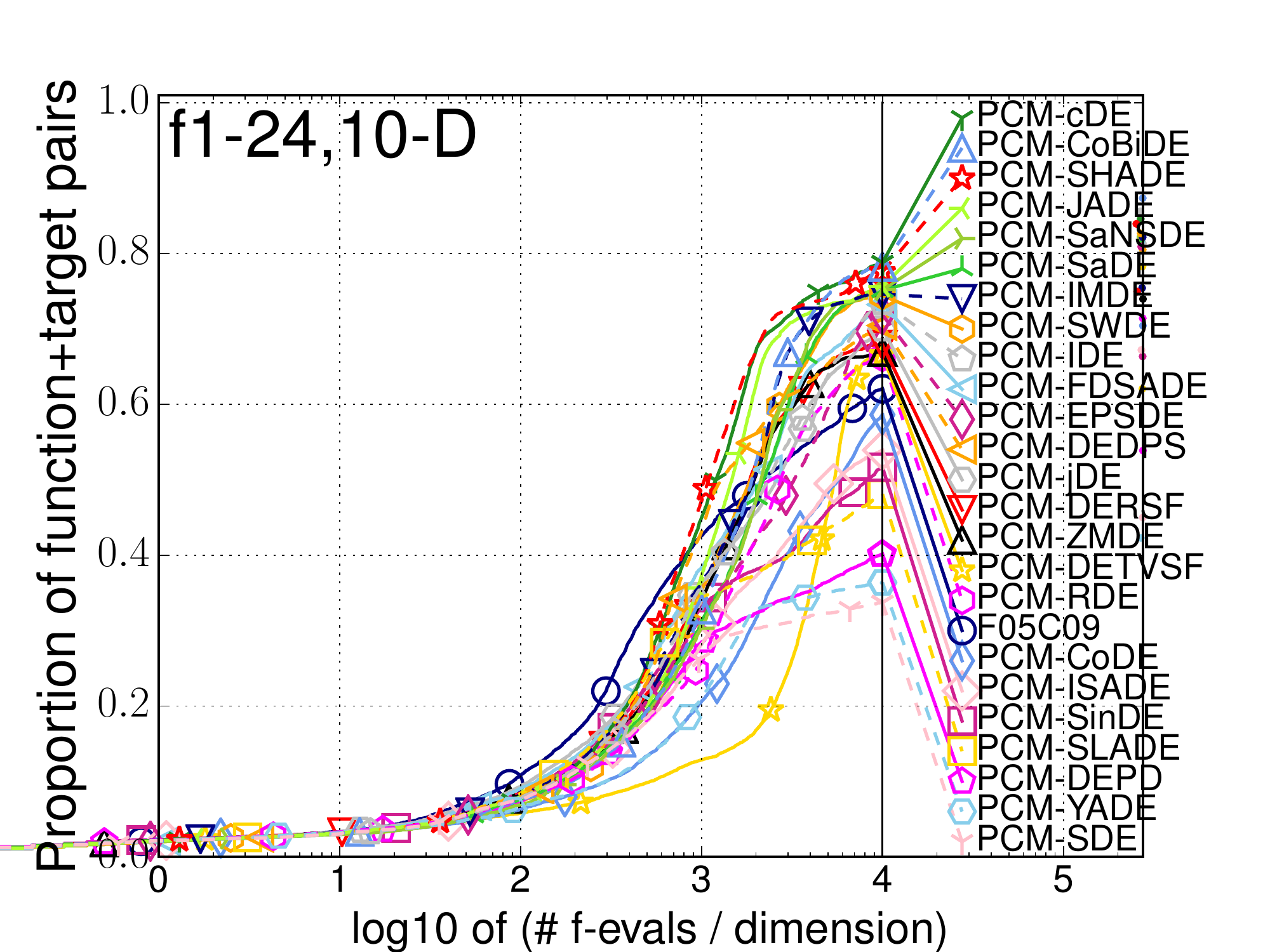}
}
\caption{
\small
Comparisons of the 25 methods (the 24 PCMs and $F05\CR09$) with the rand/1/bin and current-to-$p$best/1/bin operators on the 10-dimensional BBOB benchmark set  (higher is better).
}
\label{fig:bbob_de_restart_default_settings_bin}
  \end{center}
\end{figure}

\subsection{The performance of PCMs with default hyperparameters}
\label{sec:results-pcm-de-default}

Fig. \ref{fig:bbob_de_restart_default_settings_bin} compares the 25 methods using the rand/1/bin and current-to-$p$best/1/bin operators on the 10-dimensional BBOB benchmark set.
Fig. \ref{fig:bbob_de_restart_default_settings_bin} shows the bootstrapped Empirical Cumulative Distribution Function (ECDF) \cite{HansenABTT16} of the number of function evaluations (FEvals) divided by dimension for 51 targets in $10^{[-8..2]}$ for all 24 BBOB functions.
We used the COCO software to generate the ECDF figures.
The results of the 25 methods using all 16 operators on the 24 BBOB functions with $D \in \{2, 3, 5, 10, 20, 40\}$ are in Figs. S.1--S.16 in the supplementary file.
In this experiment, hyperparameter settings which are recommended in each paper were used for each PCM.
For details, see Tables S.2--S.26 in the supplement.

In Fig. \ref{fig:bbob_de_restart_default_settings_bin}, the vertical axis ``proportion of function $+$ target pairs'' indicates the proportion of target objective values which a given algorithm can reach within specified evaluations.
For example, in Fig.\ref{fig:bbob_de_restart_default_settings_bin}(b), PCM-CoDE reaches about $20$ percent of all target values within $1\,000 \times D$ evaluations.
If an algorithm finds the optimal solutions on all 24 functions in all 15 runs, the vertical value becomes $1$. A more detailed explanation of the ECDF is in Section S.4 in the supplement.

\noindent{\bf Statistical Analysis: }
In addition to ECDF figures, Figs. S.17--S.32 in the supplementary file show the average performance score (APS) \cite{BaderZ11} based on the error value $|f(\vector{x}^{\rm bsf} - f(\vector{x}^{*})|$ of the 25 methods for all 24 functions, where $\vector{x}^{\rm bsf}$ is the best-so-far solution found during the search process, and $\vector{x}^{*}$ is the optimal solution of a given problem.
The APS value was calculated using the Wilcoxon rank-sum test with $p < 0.05$.
For details of the APS, see Section S.5 in the supplementary file.
Since there is no significant difference between results based on the ECDF and the APS, we discuss the performance of the 25 methods based only on the ECDF figures.

{\em Recall that our interest is in the performance of PCMs, not the performance of their original, complex DE algorithms.}
Also, we are not interested in which algorithmic configuration (i.e., a particular combination of a PCM and a variation operator) performs best.
When we say below, e.g.,  ``PCM1 performs better than PCM2'', such claims are limited to the PCMs as implemented in our
standardized, experimental harness.
Thus, we make no claims as to whether the original, complex DE algorithm with PCM1 performs better than that with PCM2.


Fig. \ref{fig:bbob_de_restart_default_settings_bin}(a) shows that PCM-IDE has good performance on the 10-dimensional BBOB functions within $1\,000 \times D$ evaluations when using the rand/1/bin operator.
Beyond $2\,000 \times D$ evaluations, PCM-SHADE performs best.
Fig. \ref{fig:bbob_de_restart_default_settings_bin}(b) indicates that $F05\CR09$ (the static parameter setting) outperforms all 24 PCMs within $800 \times D$ evaluations when using the current-to-$p$best/1/bin operator.
For more than $1\,000 \times D$ evaluations, PCM-cDE, PCM-CoBiDE, and PCM-SHADE perform well.
The performance of PCM-DETVSF dramatically improved at the end of the search due to its time-dependent control strategy. 
Due to space constraints, detailed results for for the other DE operators are described in Section S.6 in the supplement.



Table \ref{tab:best_pcms} summarizes the best-performing PCM for each operator on the 24 BBOB functions with each dimensionality $D$, with respect to the APS value at the end of the search.
%
%
%
Interestingly, Table \ref{tab:best_pcms}, shows that the baseline, static parameter setting ($F=0.5$ and $\CR=0.9$) is highly competitive with the 24 PCMs on low-dimensional problems ($D \in \{2, 3\}$) for most operators.
Thus, the fixed parameter setting is suitable for low-dimensional problems.
PCM-SHADE performs well for most variation operators for $D \in \{3, 5, 10, 20\}$, and PCM-CoBiDE is competitive with PCM-SHADE.
For $D\geq 20$, PCM-cDE is the most suitable for five DE operators.
Also, PCM-ISADE, PCM-JADE, and PCM-IDE have the best performance for a particular operator for $D=40$.
These observations suggest that PCMs other than PCM-SHADE are likely to work well for high-dimensional problems.
In addition, Table \ref{tab:best_pcms} indicates that 
there are certain combinations of PCM and DE operator which tend to perform particularly well together.
For example, for the best/1/sec operator, PCM-CoBiDE has the best performance for all $D \geq 5$. 


\begin{table}
\begin{center}
  \caption{\small 
Best-performing  PCM for each DE operator on the BBOB problems (summary data; see Supplementary File for details). 
For each dimensionality $D$ and for each PCM, we list the cases where the combination of that PCM 
and a DE operator $O$ performed best (among all combinations of $O$ with the all tested PCMs).
The numbers stand for: (1) rand/1, (2), rand/2, (3) best/1, (4) best/2, (5) current-to-rand/1, (6) current-to-best/1, (7) current-to-$p$best/1, and (8) rand-to-$p$best/1. Also, (b) and (s) denote binomial and shuffled exponential crossover methods, respectively. Their combinations represent the 16 DE operators, e.g., (3s) indicates the best/1/sec operator.
For example, for $D=10$, for both the current-to-$p$best/1/best/sec and rand-to-$p$best/1/sec operators,
PCM-cDE performs best.
}
{\scriptsize
  \label{tab:best_pcms}
\scalebox{0.94}[1]{ 
\begin{tabular}{lccccccccc}
\midrule
 & $D=2$ & $D=3$ & $D=5$ & $D=10$ & $D=20$ & $D=40$\\ 
\toprule
%
\raisebox{5em}{$F05C09$} & \shortstack[l]{1b, 2b,\\3b, 6b,\\7b, 8b,\\ 1s, 3s,\\ 4s, 8s} & \raisebox{2.5em}{\shortstack[l]{4b, 8b,\\ 2s, 4s,\\7s, 8s}} & \raisebox{5em}{4s} & &  & \\
\midrule
PCM-SinDE & 2s, 7s & & & & 2s & \\
\midrule
\raisebox{2.25em}{PCM-cDE} & \raisebox{1.25em}{\shortstack[l]{4b, 5b,\\5s, 6s}} & & \raisebox{2.5em}{8b} & \raisebox{2.5em}{7s, 8s} & \shortstack[l]{1s, 5s,\\7s, 8s,\\ 6s} & \shortstack[l]{5b, 7b,\\5s, 7s,\\ 8s}\\
\midrule
PCM-ISADE &  &  &  & & & 2b\\
\midrule
PCM-JADE & & & & & 6b & 6b\\
\midrule
\raisebox{6em}{PCM-SHADE} & & \raisebox{3.75em}{\shortstack[l]{1b, 2b,\\5b, 1s,\\3s, 5s}} & \raisebox{2.5em}{\shortstack[l]{1b, 2b,\\4b, 5b,\\1s, 2s,\\5s}} & \shortstack[l]{1b, 2b,\\4b, 5b,\\ 6b, 8b,\\ 1s, 2s,\\ 4s, 5s,\\ 6s} & \raisebox{2.5em}{\shortstack[l]{1b, 2b,\\3b, 4b,\\5b, 7b,\\ 8b}} & \raisebox{3.75em}{\shortstack[l]{1b, 3b,\\4b, 8b,\\6s}}\\
\midrule
\raisebox{0.5em}{PCM-CoBiDE} & & \shortstack[l]{3b, 6b,\\7b, 6s} & \shortstack[l]{3b, 7b,\\3s, 8s} & \shortstack[l]{3b, 7b,\\ 3s} & \raisebox{0.5em}{3s, 4s} & \raisebox{0.5em}{3s, 4s}\\
\midrule
\raisebox{0.5em}{PCM-IDE} & & & \shortstack[l]{6b, 6s,\\7s} & & & \raisebox{0.5em}{1s, 2s}\\
\midrule
\end{tabular}
}
}
 \end{center}
\end{table}


%
Overall, we observed that 
the performance rankings of the PCMs  
depend significantly on the dimensionality of the target problems, resource budget, and the types of variation operators used.
%
The dependence of PCM performance on various factors has not been investigated in depth in the literature.


\begin{observation}
The performance of the PCMs relative to each other depends significantly on (1) the dimensionality of the target problems, (2) available budget (function evaluations), and (3) the types of DE operators. 
\end{observation}


Our results provide useful insights which may guide the improvement of DEs.
%
%
Cooperative co-evolution DEs that decompose a given problem into multiple lower-dimensional subproblems are popular approaches for high-dimensional continuous optimization \cite{OmidvarLMY14}.
The same DE (e.g., ``SaNSDE'' in \cite{OmidvarLMY14}) is usually applied to each decomposed subproblem in the general framework.
However, it may be better to use DEs with different PCMs for each subproblem according to their dimensionality $D$ (e.g., $F05\CR09$ for 2- and 3-dimensional subproblems and PCM-cDE for 40-dimensional subproblems).
Our results show that the best PCM depends on the three factors (see Observation 1). 
Thus, a design of a hyper-heuristic \cite{BurkeGHKOOQ13} method that adaptively selects an appropriate PCM during the search process is an interesting direction for future work.


Although it is difficult to determine a single ``best'' PCM that works well for all 16 DE operators due to the above-mentioned reason, PCM-cDE, PCM-SHADE, and PCM-CoBiDE perform well for most of the 16 variation operators on almost all problems.
While only PCM-RDE was originally designed for DE algorithms using (shuffled) exponential crossover, PCM-cDE, PCM-SHADE, and PCM-CoBiDE perform well with both binomial and shuffled exponential crossover. 
In an approximated optimal parameter adaptation process which was found in \cite{TanabeF16}, values of $F$ and $\CR$ are generated in the extreme regions $[0, 0.1]$ and $[0.9, 1]$.
PCM-cDE and PCM-CoBiDE are also capable of generating parameter values in such extreme regions.
PCM-SHADE stores a diverse set of control parameters in the historical memories $\vector{H}^F$ and $\vector{H}^{\CR}$, which potentially allow to sample parameter values in the extreme regions.
This is likely the reason why the three PCMs show good performance among the 25 methods.

\begin{observation}
PCM-cDE, PCM-SHADE, and PCM-CoBiDE perform well for most of the 16 variation operators on almost all problems when using the hyperparameter settings recommended in the literature.
\end{observation}

\subsection{The performance of PCMs with tuned hyperparameters}
\label{sec:results-pcm-de-tuned}



As described in Section \ref{sec:pcm}, most of PCMs have some hyperparameters for controlling the $F$ and $\CR$ parameters (e.g., $\tau_F$ and $\tau_{\CR}$ for PCM-jDE).
The hyperparameter settings recommended in the literature were used for each PCM in the performance comparison presented in Section \ref{sec:results-pcm-de-default}.
However, in general, the performance of PCMs depends significantly on their hyperparameter settings \cite{ZhangS09,ZamudaB15}.
Some PCMs (i.e., PCM-DERSF, PCM-DETVSF, and PCM-DEPD) also require a static, 
fixed $\CR$ value.
Although the population size $N$ was set to $5 \times D$ for all methods in Section \ref{sec:results-pcm-de-default}, a suitable $N$ value differs from each method \cite{Piotrowski17}.
Thus, the parameter settings recommended in the literature for each method may not be the most appropriate values for our benchmark problems.


In this section, we investigate the performance of PCMs with tuned parameter settings found by an automatic algorithm configurator.
In our benchmarking study, we used SMAC \cite{HutterHL11}, which is a surrogate-model based configurator.
SMAC can be used to tune real-valued, integer-valued, categorical, and conditional parameters.
We used the latest version of SMAC (version $2.10.03$) downloaded from the authors' website (\url{http://www.cs.ubc.ca/labs/beta/Projects/SMAC/}).

\begin{figure}[t]
\newcommand{\widthvar}{0.5}
  \begin{center} 
    \subfloat[PCMs with the default parameter settings]{
      \includegraphics[width=\widthvar\textwidth]{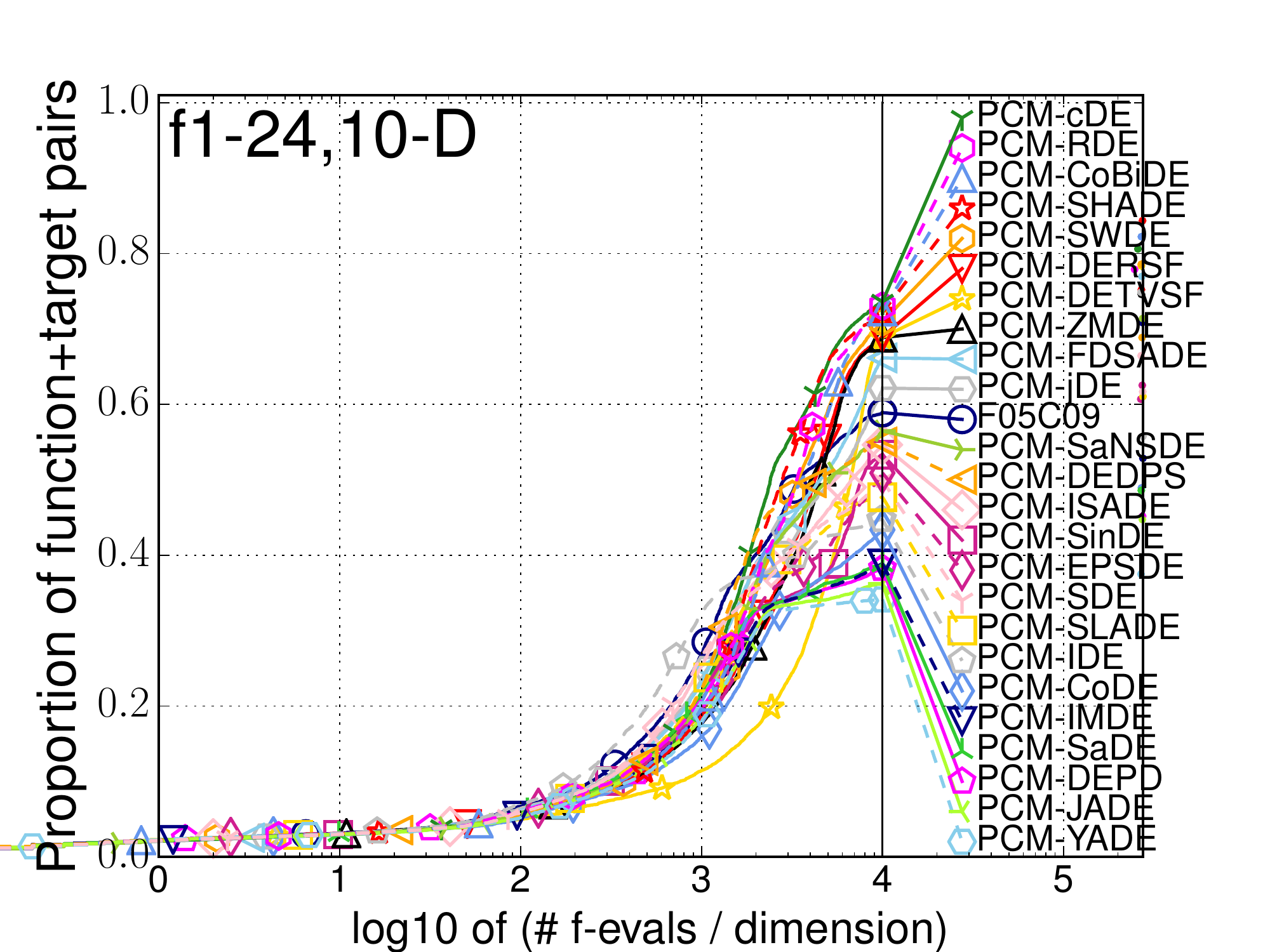}
    }
\\
    \subfloat[PCMs with the tuned parameter settings]{
      \includegraphics[width=\widthvar\textwidth]{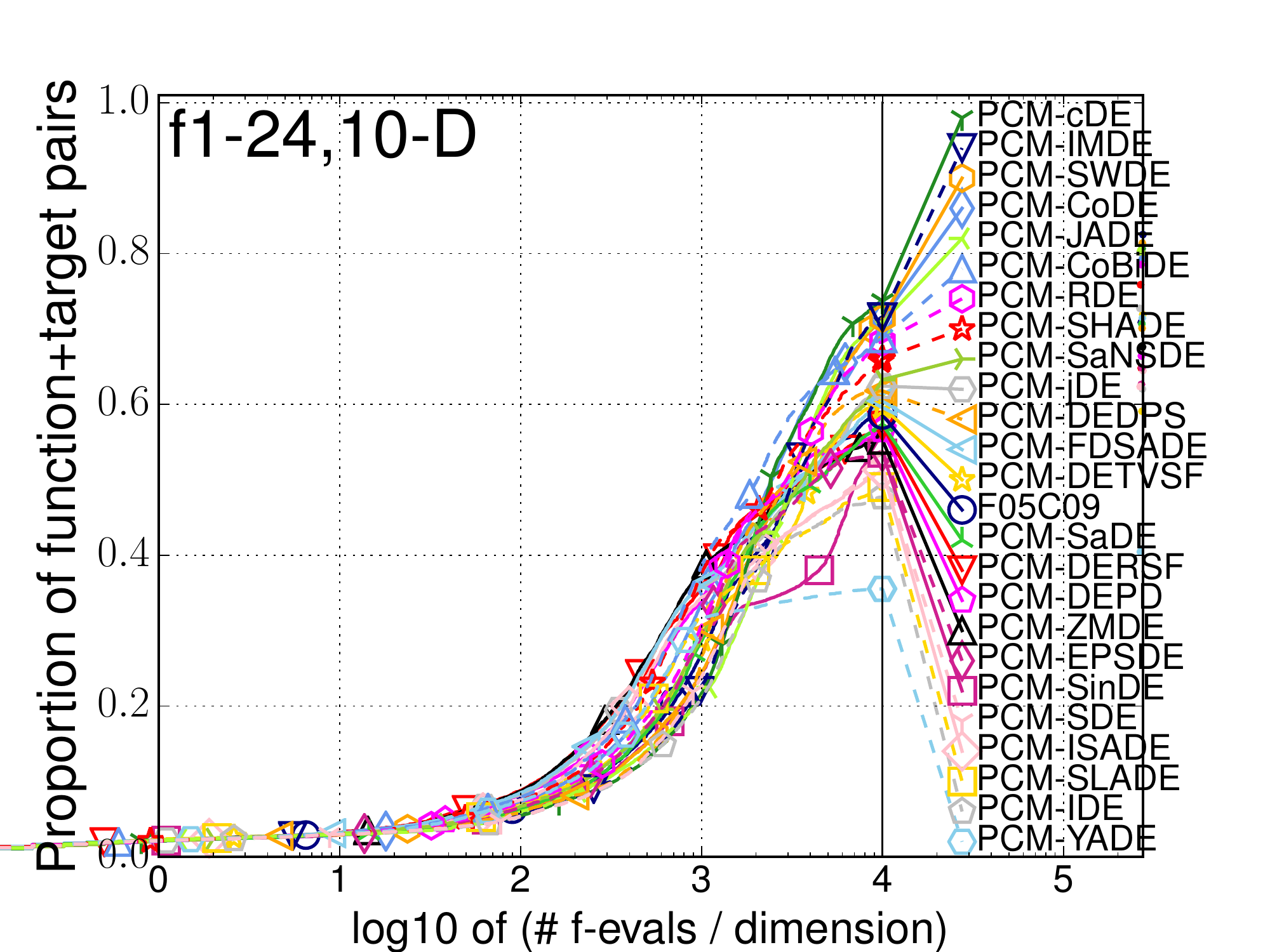}
    }
\caption{
\small
Comparisons of the 25 methods (the 24 PCMs and $F05\CR09$) using the rand/1/sec operator on the BBOB benchmarks ($D = 10$).
Figs. (a) and (b) show experimental results of the 25 methods with the default and tuned parameter configurations.
}
\label{fig:bbob_de_restart_tuned_settings}
  \end{center}
\end{figure}

The evaluation function used by SMAC to assess the quality of a candidate DE configuration was the mean of the difference between the objective value of the best-so-far solution found by the DE configuration and the optimal value for each training problem.
We used 16 functions $f^{\rm cec14}_1, ..., f^{\rm cec14}_{16}$ with $D \in \{2, 10, 20 \}$ from the CEC2014 benchmarks \cite{LiangQS14} as the training problem set.
We excluded $F^{\rm cec14}_{17}, ..., F^{\rm cec14}_{30}$ because their two-dimensional versions were not included in the CEC2014 benchmarks.
Each run of SMAC was limited to $3\,000$ DE configurations.
For each PCM as well as the static DE without any PCM, three independent SMAC runs were performed.
Then, for each method, we selected the best parameter settings from the three configurations obtained by SMAC and the recommended parameter setting.

Tables S.2--S.26 in the supplementary file show the search range of each parameter, three configurations found by SMAC, and the best parameter setting for each method.
Since the automated parameter configuration of the 25 methods required high computational cost, we conducted their parameter tuning only for the following four representative operators: rand/1/bin, rand/1/sec, current-to-$p$best/1/bin, and current-to-$p$best/1/sec.
The rand/1 operator is the most basic mutation strategy, and current-to-$p$best/1 \cite{ZhangS09} is one of the most commonly used scheme in recent work (e.g., \cite{IslamDGRS12,TanabeF13,WangLHL14,SarkerER14,YuSCZGLLZ14,TangDL15}).

For some PCMs, parameter settings found by SMAC differ significantly from the recommended values.
For example, for PCM-JADE (Table S.16), while its default values of $\mu_F$ and $\mu_{\CR}$ are $0.5$, the tuned values found by SMAC for the rand/1/sec operator take extreme values ($0.15$ and $1$, respectively).
However, for some PCMs (e.g., PCM-cDE as shown in Table S.13), SMAC could not find parameter settings which were better than the default settings.
The reason for this may be that 
many of the recommended parameter settings for the PCMs have already been tuned by the original authors 
on standard benchmark problems similar to those in the BBOB benchmarks, 
as well as benefitting from the community knowledge acquired in  large-scale parameter studies in such as \cite{GamperleMK02, ZielinskiWLK06, BrestGBMZ06}, making it difficult to obtain further improvement using current algorithm configurators.


Fig. \ref{fig:bbob_de_restart_tuned_settings} compares DE algorithms using the rand/1/sec operator with the 25 methods on the 10-dimensional BBOB benchmark set.
Other results can be found in Figs. S.33--S.36 in the supplementary file.
Figs. S.37--S.40 also show the APS of the 25 methods for all 24 BBOB functions.
Figs. \ref{fig:bbob_de_restart_tuned_settings}(a) and \ref{fig:bbob_de_restart_tuned_settings}(b) show experimental results for the 25 methods with the default and tuned parameter configurations, respectively.
The performance of some PCMs is improved after the parameter tuning.
For example, while PCM-JADE, PCM-IMDE, and PCM-CoDE with the default parameter settings does not work well (Fig. \ref{fig:bbob_de_restart_tuned_settings}(a)), they with the tuned configurations have good performance at the end of the search (Fig. \ref{fig:bbob_de_restart_tuned_settings}(b)).
In summary, the results indicate that the performance of some PCMs can be improved by hyperparameter tuning.

However, the results significantly depend on the variation operator used and the dimensionality of the target problems.
Also, for most of the cases, the three PCMs (PCM-cDE, PCM-SHADE, and PCM-CoBiDE), which had good performance in the experiments using recommended parameter settings (see Section \ref{sec:results-pcm-de-default}), still perform well among the 25 methods in this experiment (see Figs. S.33--S.36).

\begin{observation}
PCM-cDE, PCM-SHADE, and PCM-CoBiDE perform well for most of the four variation operators on almost all problems using parameter settings obtained using an automatic algorithm configurator.
\end{observation}

\begin{figure}[t]
\newcommand{\widthvar}{0.5}
  \begin{center} 
\includegraphics[width=\widthvar\textwidth]{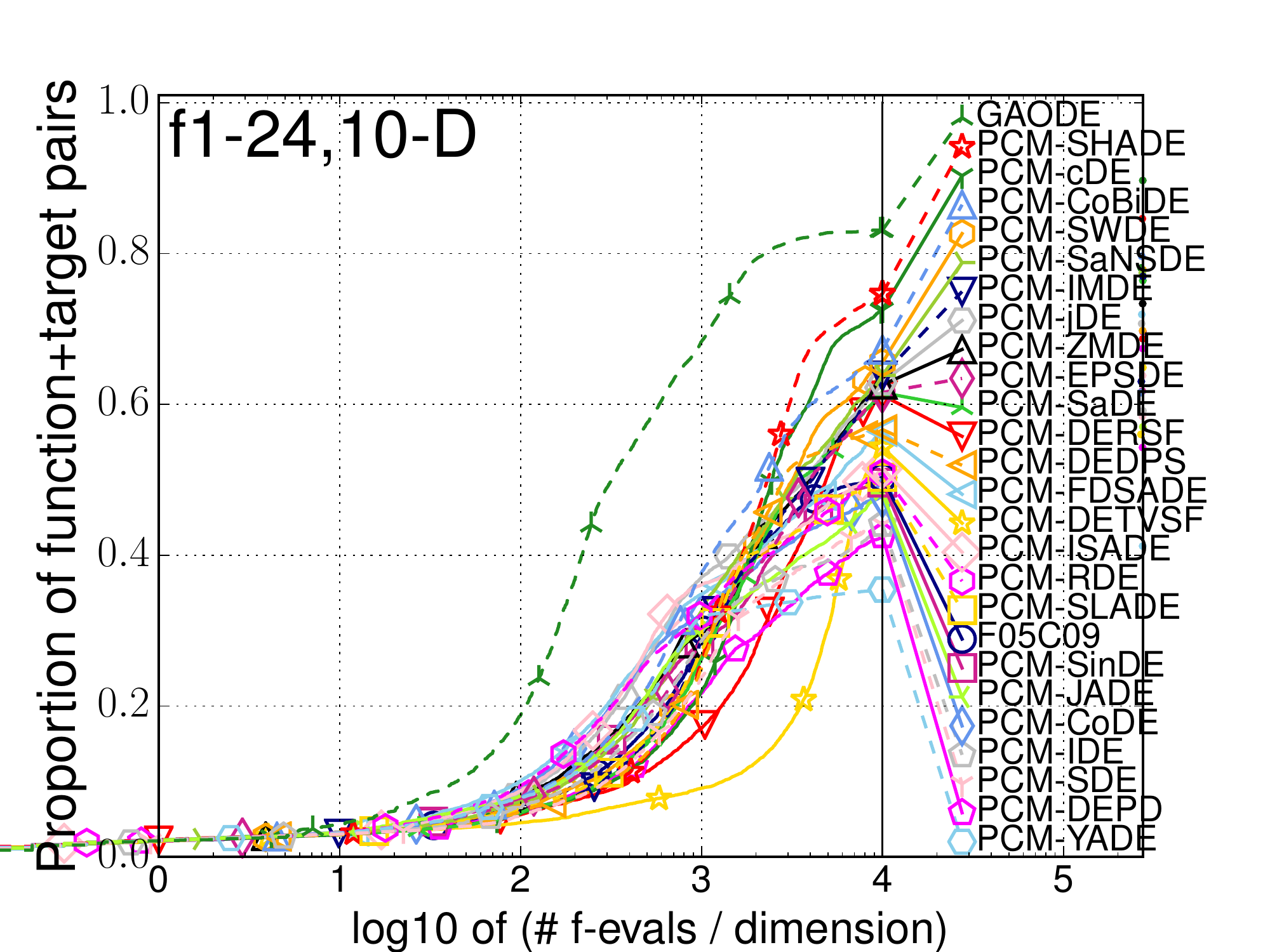}
\caption{
\small
Comparison of the 25 methods with GAODE on the BBOB benchmarks ($D =10$). 
Tuned hyperparameter settings were used for each PCM.
The rand/1/bin operator was used for all methods.
}
\label{fig:bbob_vs_gaode}
  \end{center}
\end{figure}

\subsection{Is there still room for significant improvement of PCMs?}
\label{sec:results-pcm-vs-gao}

Next, we compare the performance of existing PCMs with 
that of the Greedy Approximate Oracle method (GAO) \cite{TanabeF16},
an oracle-based method for obtaining a (lower) bound on ``optimal'' PCM behavior.
Such comparison give some indication of how much room there is for further improvement beyond the current state-of-the-art.

GAO is a simulation based method for {\em approximating} an optimal parameter adaptation process.
For each step of the search (i.e., the parameter sampling of $F_{i,t}$ and $\CR_{i,t}$ for the individual $\vector{x}^{i,t}$), GAO randomly samples many possible control parameter sets to {\em retrospectively} identify a control parameter set which would have yielded the best-expected result (with respect to 1-step-lookahead) on that step.
By repeating this process until the search terminates, GAO obtains a parameter adaptation process that is approximately optimal with respect to 1-step-lookahead.
It should be emphasized that GAO is a {\em tool} for understanding the limitation of APCMs, not an actual APCM.
Below, ``GAODE'' denotes a DE that incorporates GAO to adjust values of $F_{i,t}$ and $\CR_{i,t}$.

Fig. \ref{fig:bbob_vs_gaode} shows the comparison of the 25 methods with GAODE on the 10-dimensional BBOB functions. 
Detailed results on the 24 BBOB functions with $D \in \{2, 3, 5, 20\}$ are in Fig. S.41 in the supplementary file.
The experimental data for GAODE was derived from \cite{TanabeF16}.
Tuned hyperparameter settings were used for each PCM.
Since GAO {\em currently} works well only with the rand/1/bin operator, the comparison was conducted only for DE algorithms with the rand/1/bin.

As shown in Figs. \ref{fig:bbob_vs_gaode}, all of the 25 methods are significantly outperformed by GAODE.
For example, although PCM-SHADE performs best among the 25 methods when using rand/1/bin on the 10-dimensional problems, its performance is about 10 times worse than that of GAODE in terms of convergence speed (Fig. \ref{fig:bbob_vs_gaode}).
The results on other dimensional problems in Fig. S.41 are similar. 
Furthermore, note that GAODE is only a 1-step greedy approximation of an optimal PCM process -- a true optimal PCM process should significantly outperform GAODE.
In summary, while many PCMs have been proposed in the DE community, the comparison with GAODE indicates that there is much room for development of more efficient PCMs.

\begin{observation}
Even the best current PCMs converge at least 10 times slower than a DE which has access to an oracle that provides perfect PCM behavior.
\end{observation}

\section{Conclusion}
\label{sec:conclusion}




The contributions of this paper are threefold.
The first is an in-depth review of 24 PCMs for DE.
We extracted the PCM components from DE algorithms and precisely described in a unified framework using common terminology.
Our review provides a systematic classification and characterization of PCMs for DE, facilitating the understanding of similarities/differences between the numerous proposed PCMs.


The second contribution is a large-scale, 
benchmarking study of the 24 PCMs (and a DE with fixed parameter settings) using 16 DE operators on the BBOB benchmarks \cite{hansen2012fun}.
We investigated the performance of the PCMs with both recommended as well as tuned hyperparameter settings.
We observed that although the relative performances of the PCMs  depend significantly on the three factors (the dimensionality of problems, available budget, and the variation operator used),
PCM-cDE, PCM-SHADE, and PCM-CoBiDE exhibited consistently high performance. 

The third contribution is an assessment of how far the state-of-the-art PCMs are from an ideal PCM. We compared the 24 PCMs with the oracle-based GAODE model \cite{TanabeF16}. 
The results show that the 24 PCMs perform significantly worse than GAODE, and thus, there is still much room for the future work in the development of novel, efficient PCMs.
%



We focused on PCMs for the $F$ and $\CR$ parameters.
In addition to $F$ and $\CR$, various PCMs have applied parameter control for other aspects of the DE, such as mutation strategies (e.g., \cite{QinHS09,MallipeddiSPT11,WangCZ11}) and the population size $N$ (e.g., \cite{Teo06,BrestM08,TirronenN09,ZhabitskyZ13,TanabeF14CEC}). 
Reviewing and benchmarking these other types of PCMs is an avenue for future work.

\section*{Acknowledgment}

This work was supported by the Program for Guangdong Introducing Innovative and Entrepreneurial Teams (Grant No. 2017ZT07X386), Shenzhen Peacock Plan (Grant No. KQTD2016112514355531), the Science and Technology Innovation Committee Foundation of Shenzhen (Grant No. ZDSYS201703031748284), and the Program for University Key Laboratory of Guangdong Province (Grant No. 2017KSYS008).






\ifCLASSOPTIONcaptionsoff
  \newpage
\fi



%



\bibliography{reference}

\begin{thebibliography}{10}
\providecommand{\url}[1]{#1}
\csname url@samestyle\endcsname
\providecommand{\newblock}{\relax}
\providecommand{\bibinfo}[2]{#2}
\providecommand{\BIBentrySTDinterwordspacing}{\spaceskip=0pt\relax}
\providecommand{\BIBentryALTinterwordstretchfactor}{4}
\providecommand{\BIBentryALTinterwordspacing}{\spaceskip=\fontdimen2\font plus
\BIBentryALTinterwordstretchfactor\fontdimen3\font minus
  \fontdimen4\font\relax}
\providecommand{\BIBforeignlanguage}[2]{{%
\expandafter\ifx\csname l@#1\endcsname\relax
\typeout{** WARNING: IEEEtran.bst: No hyphenation pattern has been}%
\typeout{** loaded for the language `#1'. Using the pattern for}%
\typeout{** the default language instead.}%
\else
\language=\csname l@#1\endcsname
\fi
#2}}
\providecommand{\BIBdecl}{\relax}
\BIBdecl

\bibitem{EibenHM99}
A.~E. Eiben, R.~Hinterding, and Z.~Michalewicz, ``Parameter control in
  evolutionary algorithms,'' \emph{IEEE TEVC}, vol.~3, no.~2, pp. 124--141,
  1999.

\bibitem{KarafotiasHE15}
G.~Karafotias, M.~Hoogendoorn, and A.~E. Eiben, ``{Parameter Control in
  Evolutionary Algorithms: Trends and Challenges},'' \emph{{IEEE} TEVC},
  vol.~19, no.~2, pp. 167--187, 2015.

\bibitem{Fogarty89}
T.~C. Fogarty, ``{Varying the Probability of Mutation in the Genetic
  Algorithm},'' in \emph{ICGA}, 1989, pp. 104--109.

\bibitem{Rechenberg73}
I.~Rechenberg, \emph{Evolutionsstrategie;: Optimierung technischer Systeme nach
  Prinzipien der biologischen Evolution}.\hskip 1em plus 0.5em minus
  0.4em\relax Frommann-Holzboog, 1973.

\bibitem{BackS93}
T.~B{\"{a}}ck and H.~Schwefel, ``An overview of evolutionary algorithms for
  parameter optimization,'' \emph{Evol. Comput.}, vol.~1, no.~1, pp. 1--23,
  1993.

\bibitem{StornP97}
R.~Storn and K.~Price, ``{Differential Evolution - A Simple and Efficient
  Heuristic for Global Optimization over Continuous Spaces},'' \emph{J. Glo.
  Opt.}, vol.~11, no.~4, pp. 341--359, 1997.

\bibitem{PriceSL05}
K.~V. Price, R.~N. Storn, and J.~A. Lampinen, \emph{Differential Evolution: A
  Practical Approach to Global Optimization}, ser. Natural Computing
  Series.\hskip 1em plus 0.5em minus 0.4em\relax Springer, 2005.

\bibitem{DasS11}
S.~Das and P.~N. Suganthan, ``{Differential Evolution: A Survey of the
  State-of-the-Art},'' \emph{IEEE TEVC}, vol.~15, no.~1, pp. 4--31, 2011.

\bibitem{DasMS16}
S.~Das, S.~S. Mullick, and P.~N. Suganthan, ``Recent advances in differential
  evolution - an updated survey,'' \emph{Swarm and Evol. Comput.}, vol.~27, pp.
  1--30, 2016.

\bibitem{GamperleMK02}
R.~G\"amperle, S.~D. M\"uller, and P.~Koumoutsakos, ``A parameter study for
  differential evolution,'' in \emph{Int. Conf. on Adv. in Intelligent Systems,
  Fuzzy Systems, Evol. Comput.}, 2002, pp. 293--298.

\bibitem{ZielinskiWLK06}
K.~Zielinski, P.~Weitkemper, R.~Laur, and K.~D. Kammeyer, ``Parameter study for
  differential evolution using a power allocation problem including
  interference cancellation,'' in \emph{IEEE CEC}, 2006, pp. 1857--1864.

\bibitem{BrestGBMZ06}
J.~Brest, S.~Greiner, B.~Bo\v{s}kovi\'c, M.~Mernik, and V.~\v{Z}umer,
  ``{Self-Adapting Control Parameters in Differential Evolution: A Comparative
  Study on Numerical Benchmark Problems},'' \emph{IEEE TEVC}, vol.~10, no.~6,
  pp. 646--657, 2006.

\bibitem{NeriT10}
F.~Neri and V.~Tirronen, ``Recent advances in differential evolution: a survey
  and experimental analysis,'' \emph{Art. Intell. Rev.}, vol.~33, no. 1-2, pp.
  61--106, 2010.

\bibitem{DragoiD16}
E.~N. Dragoi and V.~Dafinescu, ``Parameter control and hybridization techniques
  in differential evolution: a survey,'' \emph{Artif. Intell. Rev.}, vol.~45,
  no.~4, pp. 447--470, 2016.

\bibitem{TanabeF14CEC}
R.~Tanabe and A.~S. Fukunaga, ``Improving the search performance of {SHADE}
  using linear population size reduction,'' in \emph{IEEE CEC}, 2014, pp.
  1658--1665.

\bibitem{ZhangS09}
J.~Zhang and A.~C. Sanderson, ``{JADE: Adaptive Differential Evolution With
  Optional External Archive},'' \emph{IEEE TEVC}, vol.~13, no.~5, pp. 945--958,
  2009.

\bibitem{WangCZ11}
Y.~Wang, Z.~Cai, and Q.~Zhang, ``{Differential Evolution With Composite Trial
  Vector Generation Strategies and Control Parameters},'' \emph{IEEE TEVC},
  vol.~15, no.~1, pp. 55--66, 2011.

\bibitem{ZielinskiWL08}
K.~Zielinski, X.~Wang, and R.~Laur, ``{Comparison of Adaptive Approaches for
  Differential Evolution},'' in \emph{PPSN}, 2008, pp. 641--650.

\bibitem{DrozdikAAT15}
M.~Drozdik, H.~E. Aguirre, Y.~Akimoto, and K.~Tanaka, ``Comparison of parameter
  control mechanisms in multi-objective differential evolution,'' in
  \emph{LION}, 2015, pp. 89--103.

\bibitem{QinHS09}
A.~K. Qin, V.~L. Huang, and P.~N. Suganthan, ``{Differential Evolution
  Algorithm With Strategy Adaptation for Global Numerical Optimization},''
  \emph{IEEE TEVC}, vol.~13, no.~2, pp. 398--417, 2009.

\bibitem{BrestM08}
J.~Brest and M.~S. Mau\u{c}ec, ``Population size reduction for the differential
  evolution algorithm,'' \emph{Appl. Intell.}, vol.~29, no.~3, pp. 228--247,
  2008.

\bibitem{MallipeddiSPT11}
R.~Mallipeddi, P.~N. Suganthan, Q.~K. Pan, and M.~F. Tasgetiren, ``Differential
  evolution algorithm with ensemble of parameters and mutation strategies,''
  \emph{Appl. Soft Comput.}, vol.~11, pp. 1679--1696, 2011.

\bibitem{ChiangCL13}
T.~Chiang, C.~Chen, and Y.~Lin, ``Parameter control mechanisms in differential
  evolution: {A} tutorial review and taxonomy,'' in \emph{IEEE SDE}, 2013, pp.
  1--8.

\bibitem{hansen2012fun}
N.~Hansen, S.~Finck, R.~Ros, and A.~Auger, ``Real-parameter black-box
  optimization benchmarking 2009: Noiseless functions definitions,'' INRIA,
  Tech. Rep., 2009.

\bibitem{HutterHL11}
F.~Hutter, F.~H. Hoos, and K.~Leyton{-}Brown, ``{Sequential Model-Based
  Optimization for General Algorithm Configuration},'' in \emph{LION}, 2011,
  pp. 507--523.

\bibitem{TanabeF16}
R.~Tanabe and A.~Fukunaga, ``{How Far Are We from an Optimal, Adaptive DE?}''
  in \emph{PPSN}, 2016, pp. 145--155.

\bibitem{BrestBGZM07}
J.~Brest, B.~Boskovic, S.~Greiner, V.~Zumer, and M.~S. Maucec, ``Performance
  comparison of self-adaptive and adaptive differential evolution algorithms,''
  \emph{Soft Comput.}, vol.~11, no.~7, pp. 617--629, 2007.

\bibitem{Tvrdik09}
J.~Tvrd{\'{\i}}k, ``Adaptation in differential evolution: {A} numerical
  comparison,'' \emph{Appl. Soft Comput.}, vol.~9, no.~3, pp. 1149--1155, 2009.

\bibitem{SeguraCSL15}
C.~Segura, C.~A.~C. Coello, E.~Segredo, and C.~Le{\'{o}}n, ``On the adaptation
  of the mutation scale factor in differential evolution,'' \emph{Opt.
  Letters}, vol.~9, no.~1, pp. 189--198, 2015.

\bibitem{SeguraCSL14}
------, ``An analysis of the automatic adaptation of the crossover rate in
  differential evolution,'' in \emph{IEEE CEC}, 2014, pp. 459--466.

\bibitem{ZhangS09_books}
J.~Zhang and A.~Sanderson, \emph{Adaptive differential evolution: a robust
  approach to multimodal problem optimization}.\hskip 1em plus 0.5em minus
  0.4em\relax Springer, 2009, vol.~1.

\bibitem{TanabeF14PPSN}
R.~Tanabe and A.~Fukunaga, ``{Reevaluating Exponential Crossover in
  Differential Evolution},'' in \emph{PPSN}, 2014, pp. 201--210.

\bibitem{IslamDGRS12}
S.~M. Islam, S.~Das, S.~Ghosh, S.~Roy, and P.~N. Suganthan, ``{An Adaptive
  Differential Evolution Algorithm With Novel Mutation and Crossover Strategies
  for Global Numerical Optimization},'' \emph{{IEEE} Trans. on SMC. B},
  vol.~42, no.~2, pp. 482--500, 2012.

\bibitem{ZouWGL13}
D.~Zou, J.~Wu, L.~Gao, and S.~Li, ``A modified differential evolution algorithm
  for unconstrained optimization problems,'' \emph{Neurocomputing}, vol. 120,
  pp. 469--481, 2013.

\bibitem{YuSCZGLLZ14}
W.~Yu, M.~Shen, W.~Chen, Z.~Zhan, Y.~Gong, Y.~Lin, O.~Liu, and J.~Zhang,
  ``{Differential Evolution With Two-Level Parameter Adaptation},''
  \emph{{IEEE} Trans. Cybernetics}, vol.~44, no.~7, pp. 1080--1099, 2014.

\bibitem{TirronenN09}
V.~Tirronen and F.~Neri, ``{Differential Evolution with Fitness Diversity
  Self-adaptation},'' in \emph{Nature-Inspired Algorithms for Optimisation},
  2009, pp. 199--234.

\bibitem{JiaGW09}
L.~Jia, W.~Gong, and H.~Wu, ``{An Improved Self-adaptive Control Parameter of
  Differential Evolution for Global Optimization},'' in \emph{ISICA}, 2009, pp.
  215--224.

\bibitem{DasKC05a}
S.~Das, A.~K., and U.~K. Chakraborty, ``Two improved differential evolution
  schemes for faster global search,'' in \emph{GECCO}, 2005, pp. 991--998.

\bibitem{DraaBB15}
A.~Draa, S.~Bouzoubia, and I.~Boukhalfa, ``A sinusoidal differential evolution
  algorithm for numerical optimisation,'' \emph{Appl. Soft Comput.}, vol.~27,
  pp. 99--126, 2015.

\bibitem{DasGM15}
S.~Das, A.~Ghosh, and S.~S. Mullick, ``{A Switched Parameter Differential
  Evolution for Large Scale Global Optimization -- Simpler May Be Better},'' in
  \emph{Mendel}, 2015, pp. 103--125.

\bibitem{AliT04}
M.~M. Ali and A.~A. T{\"o}rn, ``Population set-based global optimization
  algorithms: some modifications and numerical studies,'' \emph{Computers {\&}
  OR}, vol.~31, no.~10, pp. 1703--1725, 2004.

\bibitem{Tvrdik06}
J.~Tvrd{\i}k, ``{Competitive Differential Evolution},'' in \emph{MENDEL}, 2006,
  pp. 7--12.

\bibitem{QinS05}
A.~K. Qin and P.~N. Suganthan, ``Self-adaptive differential evolution algorithm
  for numerical optimization,'' in \emph{IEEE CEC}, 2005, pp. 1785--1791.

\bibitem{YangTY08}
Z.~Yang, K.~Tang, and X.~Yao, ``Self-adaptive differential evolution with
  neighborhood search,'' in \emph{IEEE CEC}, 2008, pp. 1110--1116.

\bibitem{YangYH08}
Z.~Yang, X.~Yao, and J.~He, ``Making a difference to differential evolution,''
  in \emph{Advances in Metaheuristics for Hard Optimization}, 2008, pp.
  397--414.

\bibitem{TanabeF13}
R.~Tanabe and A.~Fukunaga, ``{Success-History Based Parameter Adaptation for
  Differential Evolution},'' in \emph{IEEE CEC}, 2013, pp. 71--78.

\bibitem{TanabeF17}
------, ``{TPAM:} a simulation-based model for quantitatively analyzing
  parameter adaptation methods,'' in \emph{GECCO}, 2017, pp. 729--736.

\bibitem{ZhaoYHC16}
Z.~Zhao, J.~Yang, Z.~Hu, and H.~Che, ``{A differential evolution algorithm with
  self-adaptive strategy and control parameters based on symmetric Latin
  hypercube design for unconstrained optimization problems},'' \emph{EJOR},
  vol. 250, no.~1, pp. 30--45, 2016.

\bibitem{WangLHL14}
Y.~Wang, H.~Li, T.~Huang, and L.~Li, ``Differential evolution based on
  covariance matrix learning and bimodal distribution parameter setting,''
  \emph{Appl. Soft Comput.}, vol.~18, pp. 232--247, 2014.

\bibitem{SarkerER14}
R.~A. Sarker, S.~M. Elsayed, and T.~Ray, ``{Differential Evolution With Dynamic
  Parameters Selection for Optimization Problems},'' \emph{{IEEE} TEVC},
  vol.~18, no.~5, pp. 689--707, 2014.

\bibitem{TakahamaS12a}
T.~Takahama and S.~Sakai, ``{Efficient Constrained Optimization by the
  {\(\epsilon\)} Constrained Rank-Based Differential Evolution},'' in
  \emph{IEEE CEC}, 2012, pp. 1--8.

\bibitem{TangDL15}
L.~Tang, Y.~Dong, and J.~Liu, ``{Differential Evolution With an
  Individual-Dependent Mechanism},'' \emph{{IEEE} TEVC}, vol.~19, no.~4, pp.
  560--574, 2015.

\bibitem{OmranSE05}
M.~G.~H. Omran, A.~A. Salman, and A.~P. Engelbrecht, ``{Self-adaptive
  Differential Evolution},'' in \emph{CIS}, 2005, pp. 192--199.

\bibitem{KaeloA06}
P.~Kaelo and M.~M. Ali, ``A numerical study of some modified differential
  evolution algorithms,'' \emph{EJOR}, vol. 169, no.~3, pp. 1176--1184, 2006.

\bibitem{Shi99}
Y.~Shi and R.~C. Eberhart, ``Empirical study of particle swarm optimization,''
  in \emph{IEEE CEC}, vol.~3, 1999, pp. 101--106.

\bibitem{EberhartS01}
R.~C. Eberhart and Y.~Shi, ``Tracking and optimizing dynamic systems with
  particle swarms,'' in \emph{IEEE CEC}, vol.~1, 2001, pp. 94--100.

\bibitem{LiuL05}
J.~Liu and J.~Lampinen, ``A fuzzy adaptive differential evolution algorithm,''
  \emph{Soft Comput.}, vol.~9, no.~6, pp. 448--462, 2005.

\bibitem{LeeT93}
M.~A. Lee and H.~Takagi, ``{Dynamic Control of Genetic Algorithms Using Fuzzy
  Logic Techniques},'' in \emph{ICGA}, 1993, pp. 76--83.

\bibitem{Teo06}
J.~Teo, ``Exploring dynamic self-adaptive populations in differential
  evolution,'' \emph{Soft Comput.}, vol.~10, no.~8, pp. 673--686, 2006.

\bibitem{Abbass02}
H.~Abbass, ``{The self-adaptive Pareto differential evolution algorithm},'' in
  \emph{IEEE CEC}, 2002, pp. 831--836.

\bibitem{GuoLLSWC14}
H.~Guo, Y.~Li, J.~Li, H.~Sun, D.~Wang, and X.~Chen, ``Differential evolution
  improved with self-adaptive control parameters based on simulated
  annealing,'' \emph{Swarm and Evol. Comput.}, vol.~19, pp. 52--67, 2014.

\bibitem{GuoY18}
Z.~Guo and X.~Yan, ``Optimization of the p-xylene oxidation process by a
  multi-objective differential evolution algorithm with adaptive parameters
  co-derived with the population-based incremental learning algorithm,''
  \emph{Engineering Optimization}, vol.~50, no.~4, pp. 716--731, 2018.

\bibitem{DasACK09}
S.~Das, A.~Abraham, U.~K. Chakraborty, and A.~Konar, ``Differential evolution
  using a neighborhood-based mutation operator,'' \emph{IEEE TEVC}, vol.~13,
  no.~3, pp. 526--553, 2009.

\bibitem{GhoshDCG11}
A.~Ghosh, S.~Das, A.~Chowdhury, and R.~Giri, ``An improved differential
  evolution algorithm with fitness-based adaptation of the control
  parameters,'' \emph{Inf. Sci.}, vol. 181, no.~18, pp. 3749--3765, 2011.

\bibitem{NomanBI11}
N.~Noman, D.~Bollegala, and H.~Iba, ``An adaptive differential evolution
  algorithm,'' in \emph{IEEE CEC}, 2011, pp. 2229--2236.

\bibitem{IaccaCN15}
G.~Iacca, F.~Caraffini, and F.~Neri, ``{Continuous Parameter Pools in Ensemble
  Differential Evolution},'' in \emph{{IEEE} SSCI}, 2015, pp. 1529--1536.

\bibitem{MallipeddiS10SEMCCO}
R.~Mallipeddi and P.~N. Suganthan, ``{Differential Evolution Algorithm with
  Ensemble of Parameters and Mutation and Crossover Strategies},'' in
  \emph{SEMCCO}, 2010, pp. 71--78.

\bibitem{Dick10}
G.~Dick, ``The utility of scale factor adaptation in differential evolution,''
  in \emph{IEEE CEC}, 2010, pp. 1--8.

\bibitem{YangTY11}
Z.~Yang, K.~Tang, and X.~Yao, ``Scalability of generalized adaptive
  differential evolution for large-scale continuous optimization,'' \emph{Soft
  Comput.}, vol.~15, no.~11, pp. 2141--2155, 2011.

\bibitem{ChengZCN15}
J.~Cheng, G.~Zhang, F.~Caraffini, and F.~Neri, ``Multicriteria adaptive
  differential evolution for global numerical optimization,'' \emph{ICAE},
  vol.~22, no.~2, pp. 103--107, 2015.

\bibitem{FanY15}
Q.~Fan and X.~Yan, ``Self-adaptive differential evolution algorithm with
  discrete mutation control parameters,'' \emph{Expert Syst. Appl.}, vol.~42,
  no.~3, pp. 1551--1572, 2015.

\bibitem{FanY16}
------, ``{Self-Adaptive Differential Evolution Algorithm With Zoning Evolution
  of Control Parameters and Adaptive Mutation Strategies},'' \emph{{IEEE}
  Trans. Cybernetics}, vol.~46, no.~1, pp. 219--232, 2016.

\bibitem{PiotrowskiN18}
A.~P. Piotrowski and J.~J. Napiorkowski, ``{Step-by-step improvement of JADE
  and SHADE-based algorithms: Success or failure?}'' \emph{Swarm and Evol.
  Comput.}, 2018 (in press).

\bibitem{PosikK12a}
P.~Po{\v s}{\'{\i}}k and V.~Klema, ``{JADE, an adaptive differential evolution
  algorithm, benchmarked on the {BBOB} noiseless testbed},'' in \emph{GECCO
  (Companion)}, 2012, pp. 197--204.

\bibitem{TanabeF15}
R.~Tanabe and A.~Fukunaga, ``{Tuning differential evolution for cheap, medium,
  and expensive computational budgets},'' in \emph{IEEE CEC}, 2015, pp.
  2018--2025.

\bibitem{ZhabitskyZ13}
M.~Zhabitsky and E.~Zhabitskaya, ``{Asynchronous Differential Evolution with
  Adaptive Correlation Matrix},'' in \emph{GECCO}, 2013, pp. 455--462.

\bibitem{HansenABTT16}
N.~Hansen, A.~Auger, D.~Brockhoff, D.~Tusar, and T.~Tusar, ``{COCO:}
  performance assessment,'' \emph{CoRR}, vol. abs/1605.03560, 2016.

\bibitem{BaderZ11}
J.~Bader and E.~Zitzler, ``{HypE: An Algorithm for Fast Hypervolume-Based
  Many-Objective Optimization},'' \emph{Evol. Comput.}, vol.~19, no.~1, pp.
  45--76, 2011.

\bibitem{OmidvarLMY14}
M.~N. Omidvar, X.~Li, Y.~Mei, and X.~Yao, ``{Cooperative Co-Evolution With
  Differential Grouping for Large Scale Optimization},'' \emph{{IEEE} TEVC},
  vol.~18, no.~3, pp. 378--393, 2014.

\bibitem{BurkeGHKOOQ13}
E.~K. Burke, M.~Gendreau, M.~R. Hyde, G.~Kendall, G.~Ochoa, E.~{\"{O}}zcan, and
  R.~Qu, ``Hyper-heuristics: a survey of the state of the art,'' \emph{{JORS}},
  vol.~64, no.~12, pp. 1695--1724, 2013.

\bibitem{ZamudaB15}
A.~Zamuda and J.~Brest, ``Self-adaptive control parameters' randomization
  frequency and propagations in differential evolution,'' \emph{Swarm and Evol.
  Comput.}, vol.~25, pp. 72--99, 2015.

\bibitem{Piotrowski17}
A.~P. Piotrowski, ``{Review of Differential Evolution population size},''
  \emph{Swarm and Evol. Comput.}, vol.~32, pp. 1--24, 2017.

\bibitem{LiangQS14}
J.~J. Liang, B.~Y. Qu, and P.~N. Suganthan, ``{Problem Definitions and
  Evaluation Criteria for the CEC 2014 Special Session and Competition on
  Single Objective Real-Parameter Numerical Optimization},'' Zhengzhou Univ.
  and NTU., Tech. Rep., 2013.

\end{thebibliography}
\bibliographystyle{IEEEtran}

%

\begin{IEEEbiography}[{\includegraphics[width=1in,height=1.25in,keepaspectratio]{graph/tanabe.pdf}}]{Ryoji Tanabe}
  is a Research Assistant Professor with Department of Computer Science and Engineering, Southern University of Science and Technology, China.
He was a Post-Doctoral Researcher with ISAS/JAXA, Japan, from 2016 to 2017.
He received his Ph.D. in Science from The University of Tokyo, Japan, in 2016.
His research interests include stochastic single- and multi-objective optimization algorithms, parameter control in evolutionary algorithms, and automatic algorithm configuration.
  \end{IEEEbiography}


\begin{IEEEbiography}[{\includegraphics[width=1in,height=1.25in,keepaspectratio]{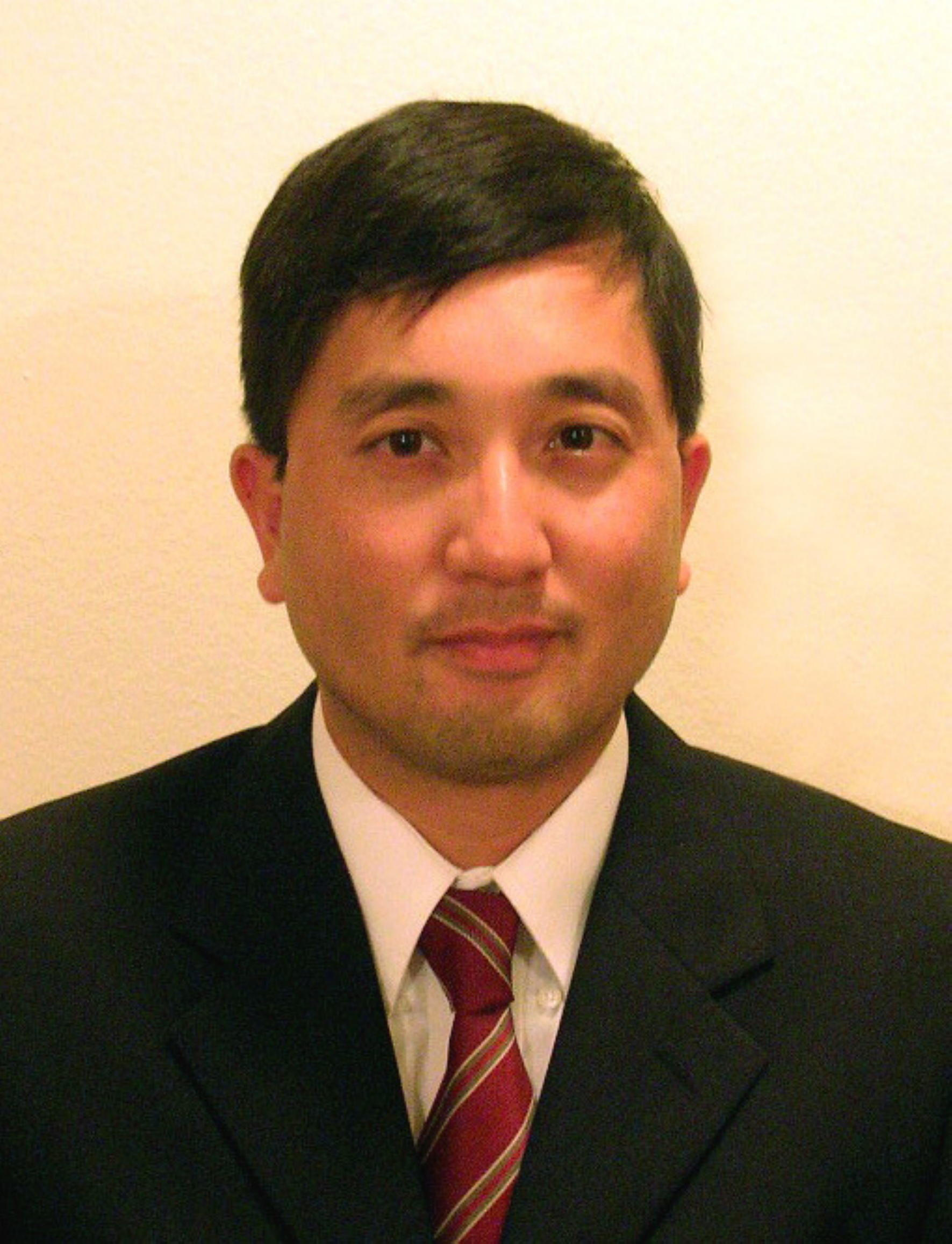}}]{Alex Fukunaga}
is a Professor in the Department of General Systems Studies, Graduate School of Arts and Sciences, in The University of Tokyo, Japan. He was previously on the faculty of the Tokyo Institute of Technology and a researcher at the NASA/Caltech Jet Propulsion Laboratory.
He received his Ph.D. in Computer Science from the University of California, Los Angeles. His research interests include heuristic search algorithms, evolutionary computation, and automated planning and scheduling.
\end{IEEEbiography}








\end{document}